\documentclass[11pt]{article}

\usepackage[dvipsnames]{xcolor}
\usepackage[acronym]{glossaries}
\usepackage{color}
\usepackage{amsmath,amsfonts,amssymb}
\usepackage{amsthm}
\usepackage{graphicx} 
\usepackage[margin=3cm]{geometry}
\usepackage{cancel}
\usepackage{verbatim}
\usepackage{url}
\usepackage{xcolor}
\usepackage{multirow}
\usepackage{caption}
\usepackage{subcaption}
\usepackage{tikz}

\newenvironment{myalign}{\par\nobreak\small\noindent\align}{\endalign}

\graphicspath{{./figs/}}

\newglossaryentry{sde}{%
  name={sde},%
  description={stochastic differential equation},%
  first={stochastic differential equation (SDE)},%
  firstplural={stochastic differential equations (SDEs)},%
  text={SDE},%
  plural={SDEs}
}

\newglossaryentry{rmse}{%
	name={rmse},%
	description={root mean squared error},%
	first={root mean squared errors (RMSE)},%
	firstplural={root mean squared errors (RMSEs)},%
	text={RMSE},%
	plural={RMSEs}
}

\newglossaryentry{nll}{%
	name={nll},%
	description={negative log loss},%
	first={negative log loss (NLL)},%
	firstplural={negative log loss (NLLs)},%
	text={NLL},%
	plural={NLLs}
}

\newglossaryentry{gp}
{%
  name={GP},%
  description={Gaussian Process},%
  first={Gaussian Process (GP)},%
  firstplural={Gaussian Processes (GPs)},%
  text={GP}%
}

\newglossaryentry{dgp}
{%
  name={DGP},%
  description={deep Gaussian Process},%
  first={deep Gaussian Process (DGP)},%
  firstplural={deep Gaussian Processes (DGPs)},%
  text={DGP}%
}

\newglossaryentry{mlp}
{%
	name={MLP},%
	description={multi layer perceptron},%
	first={multi layer perceptron (MLP)},%
	firstplural={multi layer perceptrons (MLPs)},%
	text={MLP}%
}

\newglossaryentry{sep}
{%
  name={SEP},%
  description={stochastic expectation propagation},%
  first={Stochastic Expectation Propagation (SEP)},%
  text={SEP}%
}

\newglossaryentry{ep}
{%
  name={EP},%
  description={expectation propagation},%
  first={Expectation Propagation (EP)},%
  text={EP}%
}

\newglossaryentry{gplvm}
{%
  name={GPLVM},%
  description={Gaussian Process Latent Variable Model},%
  first={Gaussian Process Latent Variable Model (GPLVM)},%
  firstplural={Gaussian Process Latent Variable Models (GPLVMs)},%
  text={GPLVM}%
}

\newglossaryentry{gpssm}
{%
  name={GPSSM},%
  description={Gaussian Process State Space Model},%
  first={Gaussian Process State Space Model (GPSSM)},%
  firstplural={Gaussian Process State Space Models (GPSSMs)},%
  text={GPSSM}%
}

\newglossaryentry{fitc}%
{%
  name={FITC},%
  description={Fully Independent Training Conditional},%
  first={Fully Independent Training Conditional (FITC)},%
  text={FITC}
}

\newcommand{\norm}{\mathcal{N}}
\newcommand{\xvec}{\mathbf{x}}
\newcommand{\zvec}{\mathbf{z}}
\newcommand{\fvec}{\mathbf{f}}

\newcommand{\uvec}{\mathbf{u}}
\newcommand{\yvec}{\mathbf{y}}

\newcommand{\hvec}{\mathbf{h}}

\newcommand{\Yvec}{\mathbf{Y}}
\newcommand{\Xvec}{\mathbf{X}}

\newcommand{\Hvec}{\mathbf{H}}

\newcommand{\zero}{\mathbf{0}}

\newcommand{\dd}{\mathrm{d}}

\DeclareMathOperator{\tr}{tr}

\allowdisplaybreaks[4]

\newcommand*{\mathcolor}{}
\def\mathcolor#1#{\mathcoloraux{#1}}
\newcommand*{\mathcoloraux}[3]{%
  \protect\leavevmode
  \begingroup
    \color#1{#2}#3%
  \endgroup
}

\makeatletter
\newcommand{\mypm}{\mathbin{\mathpalette\@mypm\relax}}
\newcommand{\@mypm}[2]{\ooalign{%
		\raisebox{.1\height}{$#1+$}\cr
		\smash{\raisebox{-.6\height}{$#1-$}}\cr}}
\makeatother

\usepackage{times}
\usepackage{graphicx} 

\RequirePackage[authoryear, round]{natbib}

\usepackage{enumitem}

\usepackage{lscape}

\usepackage{enumitem}

\newcommand\myshade{85}
\colorlet{mylinkcolor}{black}
\colorlet{mycitecolor}{YellowOrange}
\colorlet{myurlcolor}{Aquamarine}
\usepackage[breaklinks]{hyperref}
\hypersetup{
  linkcolor  = mylinkcolor!\myshade!black,
  citecolor  = mycitecolor!\myshade!black,
  urlcolor   = myurlcolor!\myshade!black,
  colorlinks = true,
}
\usepackage[hyphenbreaks]{breakurl}
\usepackage[toc,page]{appendix}

\begin{document} 





\title{Deep Gaussian Processes for Regression using Approximate Expectation Propagation}
\author{
Thang D. Bui\\
University of Cambridge\\
\texttt{tdb40@cam.ac.uk}\\
\and
Daniel Hern\'andez-Lobato\\
Universidad Aut\'onoma de Madrid\\
\texttt{daniel.hernandez@uam.es}\\
\and
Yingzhen Li\\
University of Cambridge\\
\texttt{yl494@cam.ac.uk}\\
\and
Jos\'e Miguel Hern\'andez-Lobato\\
Harvard University\\
\texttt{jmhl@seas.harvard.edu}\\
\and
Richard E. Turner\\
University of Cambridge\\
\texttt{ret26@cam.ac.uk}
}

\maketitle

\begin{abstract} 
Deep Gaussian processes (DGPs) are multi-layer hierarchical generalisations of Gaussian processes (GPs) and are formally equivalent to neural networks with multiple, infinitely wide hidden layers. DGPs are nonparametric probabilistic models and as such are arguably more flexible, have a greater capacity to generalise, and provide better calibrated uncertainty estimates than alternative deep models. This paper develops a new approximate Bayesian learning scheme that enables DGPs to be applied to a range of medium to large scale regression problems for the first time. The new method uses an approximate Expectation Propagation procedure and a novel and efficient extension of the probabilistic backpropagation algorithm for learning. We evaluate the new method for non-linear regression on eleven real-world datasets, showing that it always outperforms GP regression and is almost always better than state-of-the-art deterministic and sampling-based approximate inference methods for Bayesian neural networks. As a by-product, this work provides a comprehensive analysis of six approximate Bayesian methods for training neural networks.
\end{abstract} 


\section{Introduction}
\label{sec:intro}

\glspl{gp} are powerful nonparametric distributions over continuous functions that can be used for both supervised and unsupervised learning problems \citep{rasmussen2005gpml}. In this article, we study a multi-layer hierarchical generalisation of \glspl{gp} or \glspl{dgp} \citep{damianou-lawrence:2013a} for supervised learning tasks. A \gls{gp} is equivalent to an infinitely wide neural network with single hidden layer and similarly a \gls{dgp} is a multi-layer neural network with multiple infinitely wide hidden layers \citep{neal1995bayesian}. The mapping between layers in this type of network is parameterised by a \gls{gp}, and, as a result, \glspl{dgp} retain useful properties of \glspl{gp} such as nonparametric modelling power and well-calibrated predictive uncertainty estimates. In addition, \glspl{dgp} employ a hierarchical structure of GP mappings and therefore are arguably more flexible, have a greater capacity to generalise, and are able to provide better predictive performance \citep{andreasthesis}. This family of models is attractive as it can also potentially discover layers of increasingly abstract data representations, in much the same way as their deep parametric counterparts, but it can also handle and propagate uncertainty in the hierarchy.

The addition of non-linear hidden layers can also potentially overcome practical limitations of {\it shallow} \glspl{gp}. First, modelling real-world complex datasets often requires rich, hand-designed covariance functions. \glspl{dgp} can perform input warping or dimensionality compression or expansion, and automatically learn to construct a kernel that works well for the data at hand. As a result, learning in this model provides a flexible form of Bayesian kernel design. Second, the functional mapping from inputs to outputs specified by a \gls{dgp} is non-Gaussian which is a more general and flexible modelling choice. Third, \glspl{dgp} can repair damage done by sparse approximations to the representational power of each \gls{gp} layer. For example, pseudo datapoint based approximation methods for \glspl{dgp} trade model complexity for a lower computational complexity of $\mathcal{O}(NLM^2)$ where $N$ is the number of datapoints, $L$ is the number of layers, and $M$ is the number of pseudo datapoints. This complexity scales quadratically in $M$ whereas the dependence on the number of layers $L$ is only linear. Therefore, it can be cheaper to increase the representation power of the model by adding extra layers rather than by adding more pseudo datapoints.

The focus of this paper is Bayesian learning of \glspl{dgp}, which involves inferring the posterior over the layer mappings and hyperparameter optimisation via the marginal likelihood. Unfortunately, exact Bayesian learning in this model is analytically intractable and as such approximate inference is needed. Current proposals in the literature do not scale well and have not been compared to alternative deep Bayesian models. We will first review the model and past work in Section \ref{sec:model}, and then make the following contributions:

\begin{itemize}[leftmargin=*]
\item We propose a new approximate inference scheme for \glspl{dgp} for regression, using a sparse GP approximation, a novel approximate Expectation Propagation scheme and the probabilistic backpropagation algorithm, resulting in a computationally efficient, scalable and easy to implement algorithm (Sections \ref{sec:fitc}, \ref{sec:sep} and \ref{sec:pbp}). 
\item We demonstrate the validity of our method in supervised learning tasks on various medium to large scale datasets and show that the proposed method is always better than \gls{gp} regression and is almost always better than state-of-the-art approximate inference techniques for multi-layer neural networks (Section \ref{sec:exp}).
\end{itemize}

\section{Deep Gaussian processes}
\label{sec:model}
We first review \glspl{dgp} and existing literature on approximate inference and learning for \glspl{dgp}. Suppose we have a training set comprising of $N$ $D$-dimensional input and observation pairs $(\xvec_n, y_n)$. For ease of presentation, the outputs are assumed to be real-valued scalars, but other types of data can be easily accommodated\footnote{We also discuss how to handle non-Gaussian likelihoods in the supplementary material.}. The probabilistic representation of a \gls{dgp} comprising of $L$ layers can be written as follows,
\begin{align}
	p(f_{l}|\theta_l) &= \mathcal{GP}(f_{l};\zero,\mathbf{K}_{l}), \;\; {l=1,\cdots,L}\nonumber\\
	p(\hvec_{l}|f_{l},\hvec_{l-1},\sigma_l^2) &=\prod_{n}\mathcal{N}(h_{l,n};f_{l}(h_{l-1,n}),\sigma_{l}^2), \;\; h_{1,n} = \xvec_n\nonumber\\
	p(\yvec|f_{L},\hvec_{L-1},\sigma_L^2) &=\prod_{n}\mathcal{N}(y_{n};f_{L}(h_{L-1,n}),\sigma_{L}^2)\nonumber
\end{align}
where hidden layers\footnote{Hidden variables in the intermediate layers can and will generally have multiple dimensions but we have omitted this here to lighten the notation.} are denoted $h_{l,n}$ and the functions in each layer, $f_l$. More formally, we place a zero mean \gls{gp} prior over the mapping ${f_l}$, that is, given the inputs to $f_l$ any finite set of function values are distributed under the prior according to a multivariate Gaussian $p(\fvec_l) = \mathcal{N}(\fvec;\zero, \mathbf{K}_{\mathbf{ff}})$. Note that these function values and consequently the hidden variables are not marginally normally distributed, as the inputs are random variables. When $L=1$, the model described above collapses back to \gls{gp} regression or classification. When the inputs $\{\xvec_n\}$ are unknown and random, the model becomes a \gls{dgp} latent variable model, which has been studied in \citet{damianou-lawrence:2013a}.

\begin{figure}[!t]
\begin{center}
\centerline{\includegraphics[width=0.9\columnwidth]{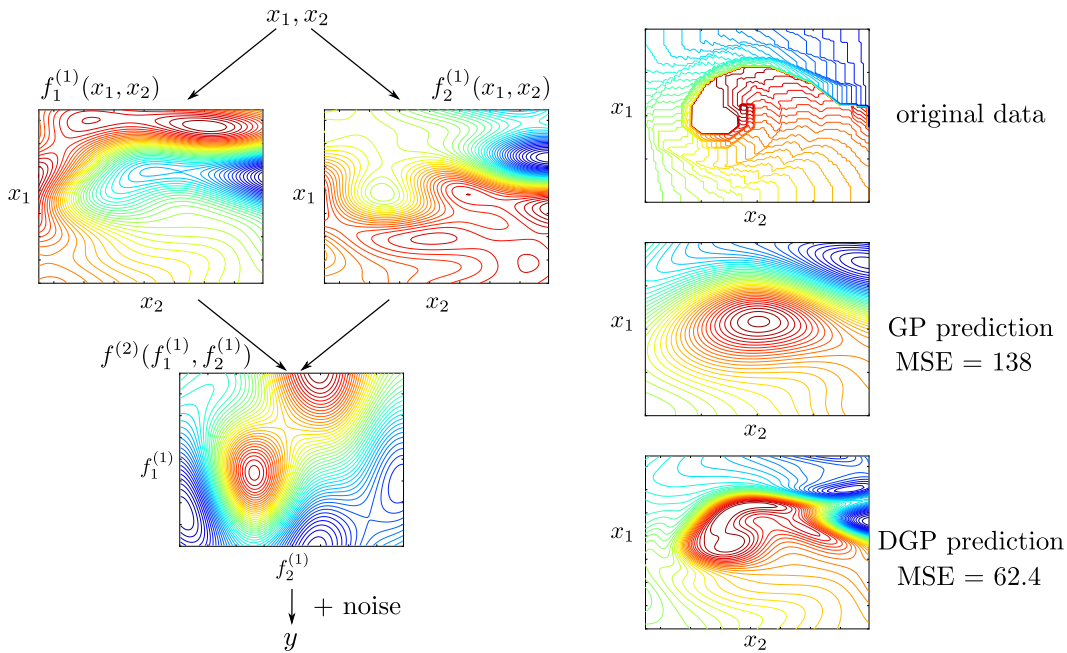}}
\caption{A deep GP example that has two GP layers and one 2-D hidden layer. The training output is the state values of the mountain car problem. The left graphs show latent functions in each layer, two functions in the first layer and one in the second layer, learnt by using the proposed approach. The right graph shows the training data [top] and the predictions of the overall function mapping from inputs to outputs made by a \gls{gp} [middle] and the \gls{dgp} on the left [bottom].}
\label{fig:two-layer-demo}
\end{center}
\end{figure}

An example of \glspl{dgp} when $L=2$ and $\mathrm{dim}(h_1) = 2$ is shown in Figure \ref{fig:two-layer-demo}. We use this network with the proposed approximation and training algorithm to fit a value function of the mountain car problem \citep{Sutton:1998} from a small number of noisy evaluations. This function is particularly difficult for models such as \gls{gp} regression with a standard exponentiated quadratic kernel due to a {\it steep value function cliff}, but is well handled by a \gls{dgp} with only two \gls{gp} layers. Interestingly the functions in the first layer are fairly simple and learn to cover or explain different parts of the input space. 

We are interested in inferring the posterior distribution over the latent function mappings and the intermediate hidden variables, as well as obtaining a marginal likelihood estimate for hyperparameter tuning and model comparison. Due to the nonlinearity in the hierarchy, these quantities are analytically intractable. As such, approximate inference is required. The simplest approach is to obtain the {\it maximum a posteriori} estimate of the hidden variables \citep{lawrence+moore:2007}. However, this procedure is prone to over-fitting and does not provide uncertainty estimates. An alternative existing approach is based on a variational-free-energy method proposed by \citet{damianou-lawrence:2013a}, extending the seminal work on variational sparse \glspl{gp} by \citet{titsias2009variational}. In this scheme, a variational approximation over both latent functions and hidden variables is chosen such that a free energy is both computationally and analytically tractable. Critically, as a variational distribution over the hidden variables is used in this approach, in addition to one over the inducing outputs, the number of variational parameters increases linearly with the number of training datapoints which hinders the use of this method for large scale datasets. Furthermore, initialisation for this scheme is a known issue, even for a modest number of datapoints \citep{turner-and-sahani:2011a}. An extension of \citet{damianou-lawrence:2013a} that has skip links from the inputs to every hidden layer in the network was proposed in \citet{dai2015variational}, based on suggestions provided in \citet{DuvRipAdaGha14}. Recent work by \citet{hensman+lawrence:2014} introduces a nested variational scheme that only requires a variational distribution over the inducing outputs, removing the parameter scaling problem of \citet{damianou-lawrence:2013a}. However, both approaches of \citet{dai2015variational} and \citet{hensman+lawrence:2014} have not been fully evaluated on benchmark supervised learning tasks or on medium to large scale datasets, nor compared to alternative deep models. 

A special case of \glspl{dgp} when $L=2$ and the sole hidden layer $h_1$ is only one dimensional is warped \glspl{gp} \citep{SneRasGha04, lazaro2012bayesian}. In \citet{lazaro2012bayesian} a variational approach, in a similar spirit to \citet{titsias2009variational} and \citet{damianou-lawrence:2013a}, was used to jointly learn the latent functions. In contrast, the latent function in the second layer is assumed to be deterministic and parameterised by a small set of parameters in \citet{SneRasGha04}, which can be learnt by maximising the analytically tractable marginal likelihood. However, the performance of warped \glspl{gp} is often similar to a standard \gls{gp}, most likely due to the narrow bottleneck in the hidden layer.

Our work differs substantially from the above and introduces an alternative approximate inference scheme for \glspl{dgp} based on three approximations. First, in order to sidestep the cubic computational cost of GPs we leverage a well-known pseudo point sparse approximation \citep{snelson+ghahramani:2006, quinonero+rasmussen:2005}. Second, an approximation to the \gls{ep} energy function \citep{seeger2005ep}, a marginal likelihood estimate, is optimised directly to find an approximate posterior over the inducing outputs. Third, the optimisation demands analytically intractable moments that are approximated by nesting Assumed Density Filtering \citep{hernandez+adams:2015}. The proposed algorithm is not restricted to the warped \gls{gp} case and is applicable to non-Gaussian observation models. 

The complexity of our method is similar to that of the variational approach proposed in \citet{damianou-lawrence:2013a}, $\mathcal{O}(NLM^2)$, but is much less memory intensive, $\mathcal{O}(LM^2)$ vs.~ $\mathcal{O}(NL + LM^2)$. These costs are competitive to those of the nested variational approach in \citet{hensman+lawrence:2014}. 


\section{The Fully Independent Training Conditional approximation}
\label{sec:fitc}
\begin{figure}[!t]
\begin{center}
\centerline{\includegraphics[width=0.5\columnwidth]{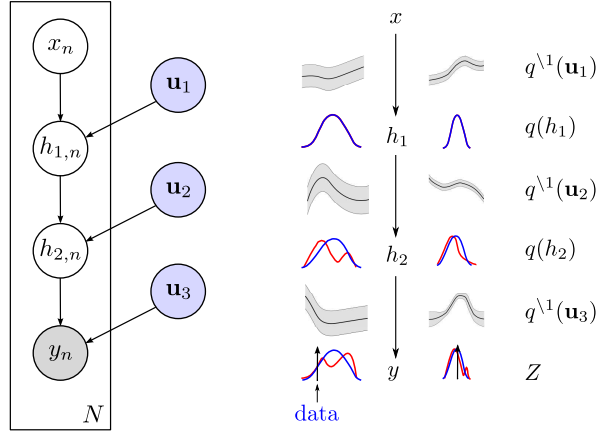}}
\caption{Left: The graphical model of our DGP-FITC model where the inducing outputs $\{\uvec_l\}$ play a role of global parameters. Right: A Gaussian moment-matching procedure to compute $\log \mathcal{Z}$. The bottom arrows denote the value of the observation and the left and right graphs [before and after an update respectively] show how the algorithm makes the final propagated Gaussian fit to the data, i.e.~the model is trained so that training points are more probable after each update. The red curves show the distribution over hidden variables before being approximated by a Gaussian in blue. Best viewed in colour.}
\label{fig:graphical-model}
\end{center}
\end{figure}

The computational complexity of full \gls{gp} models scales cubically with the number of training instances, making it intractable in practice. Sparse approximation techniques are therefore often necessary. They can be coarsely put into two classes: ones that explicitly sparsify and create a semi-parametric representation that approximates the original model, and ones that retain the original nonparametric properties and perform sparse approximation to the exact posterior. The method used here, \gls{fitc}, falls into the first category. The \gls{fitc} approximation is formed by considering a small set of $M$ function values $\uvec$ in the infinite dimensional vector $f$ and assuming conditional independence between the remaining values given the set $\uvec$ \citep{snelson+ghahramani:2006,quinonero+rasmussen:2005}. This set is often called inducing outputs or pseudo targets and their input locations $\zvec$ can be chosen by optimising the approximate marginal likelihood. The resulting model can be written as follows,
\begin{align}
	p(\uvec_{l}|\theta_l) &= \norm(\uvec_{l};\zero,\mathbf{K}_{\uvec_{l-1},\uvec_{l-1}}), \;\; {l=1,\cdots,L}\nonumber\\
	p(\hvec_{l}|\uvec_{l},\hvec_{l-1},\sigma_l^2) &=\prod_{n}\mathcal{N}(h_{l,n}; \mathbf{C}_{n,l} \uvec_{l}, \mathbf{R}_{n,l}),\nonumber\\
	p(\yvec|\uvec_{L},\Hvec_{L-1},\sigma_L^2) &=\prod_{n}\mathcal{N}(y_{n}; \mathbf{C}_{n,L} \uvec_{L}, \mathbf{R}_{n,L})\nonumber.
\end{align}
where $\mathbf{C}_{n,l} = \mathbf{K}_{\hvec_{l,n},\uvec_{l}} \mathbf{K}_{\uvec_{l},\uvec_{l}}^{-1}$ and $\mathbf{R}_{n,l} = \mathbf{K}_{\hvec_{l,n},\hvec_{l,n}} - \mathbf{K}_{\hvec_{l,n},\uvec_{l}} \mathbf{K}_{\uvec_{l},\uvec_{l}}^{-1} \mathbf{K}_{\uvec_{l},\hvec_{l,n}} + \sigma_l^2\mathrm{I}$. Note that the function outputs index the covariance matrices, for example $\mathbf{K}_{\hvec_{l,n},\uvec_{l}}$ denotes the covariance between $\hvec_{l,n}$ and $\uvec_l$, and takes $\hvec_{l-1,n}$ and $\zvec_l$ as inputs respectively. This is important when propagating uncertainty through the network. The \gls{fitc} approximation creates a semi-parametric model, but one which is cleverly structured so that the induced non-stationary noise captures the uncertainty introduced from the sparsification. The computational complexity of inference and hyperparameter tuning in this approximate model is $\mathcal{O}(NM^2)$ which means $M$ needs to be smaller than $N$ to provide any computational gain (i.e.~the approximation should be sparse). The quality of the approximation largely depends on the number of inducing outputs $M$ and the complexity of the underlying function, i.e.~if the function's characteristic lengthscale is small, $M$ needs to be large and vice versa. As $M$ tends to $N$ and $\zvec = \Xvec$, i.e.~the inducing inputs and training inputs are shared, the approximate model reverts back to the original GP model. The graphical model is shown in Figure \ref{fig:graphical-model} [left].

\section{Approximate Bayesian inference via EP}
\label{sec:sep}
Having specified a probabilistic model for data using a deep sparse Gaussian processes we now consider inference for the inducing outputs $\uvec = \{\uvec_l\}_{l=1}^L$ and learning of the model parameters $\alpha = \{\zvec_l, \theta_l \}_{l=1}^L$. The posterior distribution over the inducing outputs can be written as $p(\uvec|\Xvec,\yvec) \propto p(\uvec) \prod_n p(y_n|\uvec,\Xvec_n)$. This quantity can then be used for output prediction given a test input, $p(y^*|\xvec^*,\Xvec,\yvec) = \int \dd\uvec\, p(\uvec|\Xvec,\yvec)\, p(y^*|\uvec,\xvec^*)$. The model hyperparameters can be tuned by maximising the marginal likelihood $p(\yvec|\alpha) = \int \dd\uvec \, \dd \hvec \, p(\uvec, \hvec)\, p(\yvec|\uvec,\hvec,\alpha)$. However, both the posterior of $\uvec$ and the marginal likelihood are not analytically tractable when there is more than one \gls{gp} layer in the model. As such, approximate inference is needed; here we make use of the \gls{ep} energy function with a tied factor constraint similar to that proposed in the \gls{sep} algorithm \citep{li+etal:2015} to produce a scalable, convergent approximate inference method. 


\subsection{EP, Stochastic EP and the EP approximate energy}
In \gls{ep} \citep{minka2001ep}, the approximate posterior is assumed to be $q(\uvec) \propto p(\uvec) \prod_n \tilde{t}_n(\uvec)$ where $\{\tilde{t}_n(\uvec)\}_{n=1}^{N}$ are the approximate data factors. Each factor approximately captures the contribution of datapoint $n$ makes to the posterior and, in this work, they take an unnormalised Gaussian form. The factors can be found by running an iterative procedure which often requires several passes through the training set for convergence\footnote{We summarise the EP steps in the supplementary material.}. The \gls{ep} algorithm also provides an approximation to the marginal likelihood,
\begin{align}
\log p(\yvec | \alpha) \approx \mathcal{F} (\alpha) &= \phi(\theta) - \phi(\theta_{\mathrm{prior}}) + \sum_{n=1}^{N} \log \tilde{\mathcal{Z}}_n \nonumber\\
\text{where } \quad \log \tilde{\mathcal{Z}}_n &= \log \mathcal{Z}_n + \phi(\theta^{\setminus n}) - \phi(\theta),\nonumber
\end{align}
%
%
where $\theta, \theta^{\setminus n}$ and $\theta_\mathrm{prior}$ are the natural parameters of $q(\uvec)$, the cavity $q^{\setminus n}(\uvec)$ [$q^{\setminus n}(\uvec) \propto q(\uvec) / \tilde{t}_n(\uvec)$] and $p(\uvec)$ respectively, $\phi(\theta)$ is the log normaliser of a Gaussian distribution with natural parameters $\theta$, and $\log \mathcal{Z}_n = \log \int \dd \uvec \, q^{\setminus n}(\uvec) \, p(y_n|\uvec,\Xvec_n)$ \citep{seeger2005ep}. Unfortunately, EP is not guaranteed to converge, but if it does, the fixed points lie at the stationary points of the EP energy, which is given by $-\mathcal{F} (\alpha)$. Furthermore, EP requires the approximate factors to be stored in memory, which has a cost of $\mathcal{O}(NLM^2)$ in this application as we need to store the mean and the covariance matrix for each factor.

\subsection{Direct EP energy minimisation with a tied factor constraint}
\label{sec:ep_energy}
In order to reduce the expensive memory footprint of \gls{ep}, the data-factors are tied. That is the posterior $p(\uvec|\Xvec, \yvec)$ is approximated by $q(\uvec) \propto p(\uvec) g(\uvec)^{N}$, where the factor $g(\uvec)$ could be thought of as an {\it average} data factor that captures the average effect of a likelihood term on the posterior. Approximations of this form were recently used in the \gls{sep} algorithm \citep{li+etal:2015} and although seemingly limited, in practice were found to perform almost as well as full \gls{ep} while significantly reducing EP's memory requirement, from $\mathcal{O}(NLM^2)$ to $\mathcal{O}(LM^2)$ in our case.

The original SEP work devised modified versions of the EP updates appropriate for the new form of the approximate posterior. Originally we applied this method to DGPs (details of this approach including hyperparameter optimisation are included in the supplementary material). However, an alternative approach was found to have superior performance, which is to optimise the EP energy function directly (for both the approximating factors and the hyperparameters).  Normally, optimisation of the EP energy requires a double-loop algorithm, which is computationally inefficient, however the use of tied factors simplifies the approximate marginal likelihood and allows direct optimisation. The energy becomes,
\begin{align} 
\mathcal{F}(\alpha) 
&= \phi(\theta) - \phi(\theta_{\mathrm{prior}}) + \sum_{n=1}^{N} \left[ \log \mathcal{Z}_n + \phi(\theta^{\setminus 1}) - \phi(\theta) \right] \nonumber\\
&= (1-N) \phi(\theta) + N \phi(\theta^{\setminus 1}) - \phi(\theta_{\mathrm{prior}}) + \sum_{n=1}^{N} \log \mathcal{Z}_n \nonumber
\end{align}
since the cavity distribution $q^{\setminus n}(\uvec) \propto q(\uvec) / \tilde{t}_n(\uvec) = q(\uvec) / g(\uvec) = q^{\setminus 1}(\uvec)$ is the same for all training points. This elegantly removes the need for a double-loop algorithm, since we can posit a form for the approximate posterior and optimise the above approximate marginal likelihood directly. However, it is important to note that, in general, optimising this objective will not give the same solution as optimising the full EP energy. The new energy produces an approximation formed by averaging the moments of $q^{\setminus 1}(\uvec) \, p(y_n|\uvec,\xvec_n)$ over datapoints, whereas EP averages natural parameters, which is arguably more sensible but less tractable.

In detail, we assume the tied factor takes a Gaussian form with natural parameters $\theta_1$. As a result, the approximate posterior and the cavity are also Gaussian with natural parameters $\theta = \theta_\mathrm{prior} + N \theta_1$ and $\theta^{\setminus 1} = \theta_\mathrm{prior} + (N-1)\theta_1$ respectively. This means that we can compute the first three terms in the energy function exactly. However, it remains to compute $\log \mathcal{Z}_n = \log \int \dd \uvec \, q^{\setminus 1}(\uvec) \, p(y_n|\uvec,\xvec_n) $ which we will discuss next.

\section{Probabilistic backpropagation for deep Gaussian processes}
\label{sec:pbp}
Computing $\log \mathcal{Z}_n$ in the objective function above is analytically intractable for $L\geq1$ since the likelihood given the inducing outputs $\uvec$ is nonlinear and the propagation of the Gaussian cavity through each layer results in a complex distribution. However, for certain choices of covariance functions $\{ \mathbf{K}_{l} \}_{l=1}^L$, it is possible to use an efficient and accurate approximation which propagates a Gaussian through the first layer of the network and projects this non-Gaussian distribution back to a moment matched Gaussian before propagating through the next layer and repeating the same steps. This scheme is an algorithmic identical to Assumed Density Filtering and a central part of the probabilistic backpropagation algorithm that has been applied to standard neural networks \citep{hernandez+adams:2015}.

The aim is to compute $\log \mathcal{Z}$ and its gradients with respect to the parameters such as $\theta_1$ or the hyperparameters of the model\footnote{We ignore the data index here to lighten the notation}. By reintroducing the hidden variables in the middle layers, we perform a Gaussian approximation to $\mathcal{Z}$ in a sequential fashion, as illustrated in Figure \ref{fig:graphical-model} [right]. We take a two layer case as a running example:
\begin{align}
\mathcal{Z} &= \int \dd \uvec \, p(y|\xvec, \uvec) \, q^{\setminus 1}(\uvec) \nonumber \\
&= \int \dd h_1 \, \dd \uvec_2 \, p(y|h_1, \uvec_2) \, q^{\setminus 1}(\uvec_2) \int \dd \uvec_1 \, p(h_1|\xvec, \uvec_1) \, q^{\setminus 1}(\uvec_1) \nonumber
\end{align}
One key difference between our approach and the variational free energy method of \citet{damianou-lawrence:2013a} is that our algorithm does not retain an explicit approximate distribution over the hidden variables. Instead, we approximately integrate them out when computing $\log \mathcal{Z}$ as follows.

First, we can exactly marginalise out the inducing outputs for each \gls{gp} layer, leading to $\mathcal{Z} = \int \dd h_1 \, q(y|h_1) \, q(h_1)$ where $q(h_1) = \norm(h_1; m_1, v_1) $, $q(y|h_1) = \norm(y|h_1; m_{2|h_1}, v_{2|h_1})$ and 
\begin{align}
m_1 &= \mathbf{K}_{h_1,\uvec_1} \mathbf{K}_{\uvec_1,\uvec_1}^{-1} \mathbf{m}_1^{\setminus 1}, \nonumber\\ 
v_1 &= \sigma_1^2 + K_{h_1,h_1} - \mathbf{K}_{h_1,\uvec_1} \mathbf{K}_{\uvec_1,\uvec_1}^{-1} \mathbf{K}_{\uvec_1,h_1} + \mathbf{K}_{h_1,\uvec_1} \mathbf{K}_{\uvec_1,\uvec_1}^{-1} \mathbf{V}_1^{\setminus 1} \mathbf{K}_{\uvec_1,\uvec_1}^{-1} \mathbf{K}_{\uvec_1,h_1},\nonumber\\
m_{2|h_1} &= \mathbf{K}_{h_2,\uvec_2} \mathbf{K}_{\uvec_2,\uvec_2}^{-1} \mathbf{m}_2^{\setminus 1}, \nonumber\\
v_{2|h_1} &= \sigma_2^2 + K_{h_2,h_2} - \mathbf{K}_{h_2,\uvec_2} \mathbf{K}_{\uvec_2,\uvec_2}^{-1} \mathbf{K}_{\uvec_2, h_2} + \mathbf{K}_{h_2,\uvec_2} \mathbf{K}_{\uvec_2,\uvec_2}^{-1} \mathbf{V}_1^{\setminus 1} \mathbf{K}_{\uvec_2,\uvec_2}^{-1} \mathbf{K}_{\uvec_2,h_2}.\nonumber
\end{align}

Following \citep{GirRasQuiMur03, barber-schottky-98b, deisenroth2012ep}, we can use the law of iterated conditionals to approximate the difficult integral in the equation above by a Gaussian $\mathcal{Z} \approx \norm(y|m_2, v_2)$ where the mean and variance take the following form,
\begin{align}
m_{2} &= \mathrm{E}_{q(h_1)} [m_{2|h_1}] \nonumber \\
v_{2} &= \mathrm{E}_{q(h_1)} [v_{2|h_1}]  + \mathrm{var}_{q(h_1)}[m_{2|h_1}] \nonumber
\end{align}
which results in 
\begin{align}
m_{2} &= \mathrm{E}_{q(h_1)} [\mathbf{K}_{h_2,\uvec_2}] \mathbf{A} \nonumber\\ 
v_{2} &= \sigma_2^2 + \mathrm{E}_{q(h_1)} [K_{h_2,h_2}] + \tr \left( \mathbf{B} \mathrm{E}_{q(h_1)} [\mathbf{K}_{\uvec_2,h_2} \mathbf{K}_{h_2,\uvec_2} ] \right) - m_{2}^2\nonumber
\end{align}
where $\mathbf{A} = \mathbf{K}_{\uvec_2,\uvec_2}^{-1} \mathbf{m}_2^{\setminus 1}$ and $\mathbf{B} = \mathbf{K}_{\uvec_2,\uvec_2}^{-1} ( \mathbf{V}_2^{\setminus 1} + \mathbf{m}_2^{\setminus 1} \mathbf{m}_2^{\setminus 1, \mathrm{T}} ) \mathbf{K}_{\uvec_2,\uvec_2}^{-1} - \mathbf{K}_{\uvec_2,\uvec_2}^{-1}$. The equations above require the expectations of the kernel matrix under a Gaussian distribution over the inputs, which are analytically tractable for widely used kernels such as exponentiated quadratic, linear or a more general class of spectral mixture kernels \citep{titsias2010bayesian, wilson+adams:2013}. In addition, the approximation above can be improved for networks that have multidimensional intermediate variables, by using a Gaussian with a non-diagonal covariance matrix. We discuss this in the supplementary material.

As the mean and variance of the Gaussian approximation in each intermediate layer can be computed analytically, their gradients with respect to the mean and variance of the input distribution, as well as the parameters of the current layers are also available. Since we require the gradients of the approximation to $\log \mathcal{Z}$, we need to store these results in the forward propagation step, compute the approximate $\log \mathcal{Z}$ and its gradients at the output layer and use the chain rule in the backward step to differentiate through the ADF procedure. This procedure is reminiscent of the backpropagation algorithm in standard parametric neural networks, hence the name {\it probabilistic backpropagation} \citep{hernandez+adams:2015}.

\section{Stochastic optimisation for scalable training}
The propagation and moment-matching as described above costs $\mathcal{O}(LM^2)$ and needs to be repeated for all datapoints in the training set in batch mode, resulting in an overall complexity of $\mathcal{O}(NLM^2)$. Fortunately, the last term of the objective in Section \ref{sec:ep_energy} is a sum of independent terms, i.e.\ its computation can be distributed, resulting in a substantial decrease in computational cost. Furthermore, the objective is suitable for stochastic optimisation. In particular, an unbiased noisy estimate of the objective and its gradients can be obtained using a minibatch of training datapoints,
\begin{align}
\mathcal{F} \approx - (N-1) \phi(\theta) + N \phi(\theta^{\setminus 1}) - \phi(\theta_{\mathrm{prior}}) + \frac{N}{|B|}\sum_{b=1}^{|B|} \log \mathcal{Z}_b, \nonumber
\end{align}
where $|B|$ denotes the minibatch size.

\section{Approximate predictive distribution}
Given the approximate posterior and a new test input $x^*$, we wish to make a prediction about the test output $y^*$. That is to find $p(y^*|x^*, \Xvec, \Yvec) \approx \int\dd \uvec \, p(y^*|x^*, \uvec)\, q(\uvec|\Xvec, \Yvec)$. This predictive distribution is not analytically tractable, but fortunately, we can approximate it by a Gaussian in a similar fashion to the method described in Section \ref{sec:pbp}. That is, a single forward pass is performed, in which each layer takes in a Gaussian distribution over the input, incorporates the approximate posterior of the inducing outputs and approximates the output distribution by a Gaussian. An alternative to obtain the prediction is to forward sample from the model, but we do not use this approach in the experiments.


\section{Experiments}
\label{sec:exp}
We implement and compare the proposed approximation scheme to state-of-the-art methods for Bayesian neural networks. We first detail our implementation in Section \ref{exp:details} and then discuss the experimental results in Sections \ref{exp:uci} and \ref{exp:molecule}. 
\subsection{Experimental details} 
\label{exp:details}
In all the experiments reported here, we use Adam with the default learning rate \citep{kingma+ba:2015} for optimising our objective function. We use an exponentiated quadratic kernel with ARD lengthscales for each layer. The hyperparameters and pseudo point locations are different between functions in each layer. The lengthscales and inducing inputs of the first \gls{gp} layer are sensibly initialised based on the median distance between datapoints in the input space and the k-means cluster centers respectively. We use long lengthscales and initial inducing inputs between $[-1, 1]$ for the higher layers to force them to start with an identity mapping. We parameterise the natural parameters of the average factor and initialise them with small random values. We evaluate the predictive performance on the test set using two popular metrics: root mean squared error (RMSE) and mean log likelihood (MLL).


\subsection{Regression on UCI datasets}
\label{exp:uci}
We validate the proposed approach for training \glspl{dgp} in regression experiments using several datasets from the UCI repository. In particular, we use the ten datasets and train/test splits used in \citet{hernandez+adams:2015} and \citet{yarin+zoubin:2015}: 1 split for the {\it year} dataset [$N\approx500000, D=90$], 5 splits for the {\it protein} dataset [$\small N\approx46000, D=9$], and 20 for the others.

\begin{figure*}[!t]
\centerline{\includegraphics[width=\textwidth]{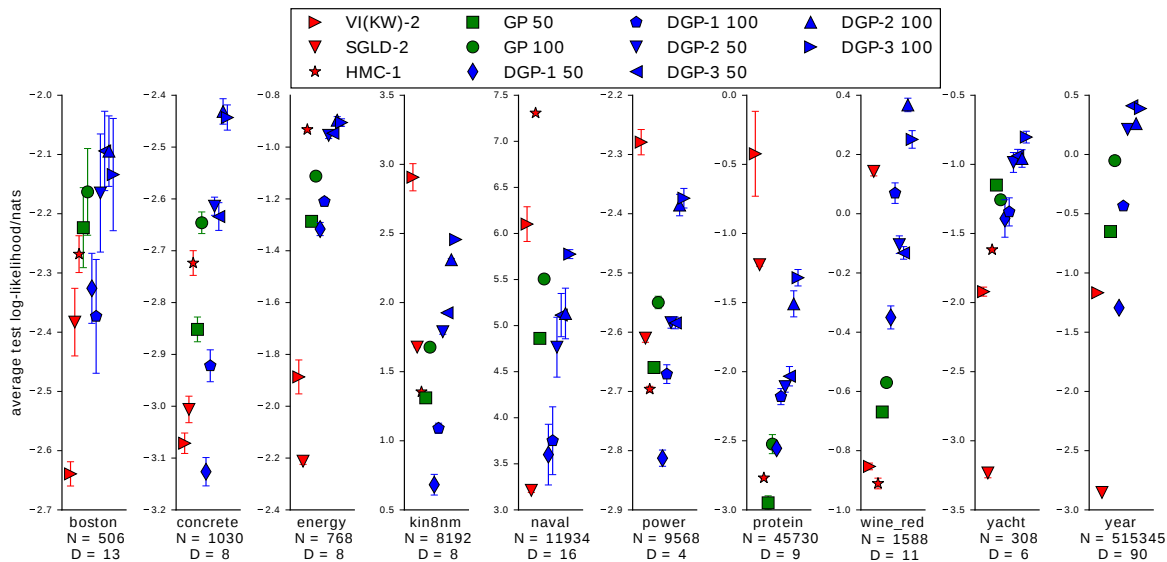}}
\caption{Average predictive log likelihood of existing approaches for BNNs and GPs, and the proposed method for DGPs, across 10 datasets. The higher the better, and best viewed in colour. Full results are included in the supplementary material.} 
\label{fig:reg_results_nll}
\end{figure*}

We compare our method (FITC-DGP) against sparse GP regression using FITC (FITC-GP) and Bayesian neural network (BNN) regression using several state-of-the-art deterministic and sampling-based approximate inference techniques. As baselines, we include the results for BNNs reported in \citet{hernandez+adams:2015}, BNN-VI(G)-1 and BNN-PBP-1, and in \citet{yarin+zoubin:2015}, BNN-Dropout-1. The results reported for these methods are for networks with one hidden layer of 50 units (100 units for {\it protein} and {\it year}). Specifically, BNN-VI(G) uses a mean-field Gaussian approximation for the weights in the network, and obtains the stochastic estimates of the bound and its gradient using a Monte Carlo approach \citep{graves2011practical}. BNN-PBP employs Assumed Density Filtering and the probabilistic backpropagation algorithm to obtain a Gaussian approximation for the weights \citep{hernandez+adams:2015}. BNN-Dropout is a recently proposed technique that employs {\it dropout} during training as well as at prediction time, that is to average over several predictions, each made by the entire network with a random proportion of the weights set to zero \citep{yarin+zoubin:2015}. We implement other methods as follows,
\begin{itemize}[leftmargin=*]
\item DGP: we evaluate three different architectures of \glspl{dgp}, each with two GP layers and one hidden layer of one, two and three dimensions respectively (DGP-1, DGP-2 and DGP-3). We include the results for two settings of the number of inducing outputs, $M=50$ and $M=100$ respectively. Note that for the bigger datasets {\it protein} and {\it year}, we use $M=100$ and $M=200$ but do not annotate this in Figure \ref{fig:reg_results_nll}. We choose these settings to ensure the run time for our method is smaller or comparable to that of other methods for BNNs.
\item GP: we use the same number of pseudo datapoints as in DGP (GP 50 and GP 100).
\item BNN-VI(KW): this method, similar to \citet{graves2011practical}, employs a mean-field Gaussian variational approximation but evaluates the variational free energy using the {\it reparameterisation trick} proposed in \citet{kingma+welling:2014}. We use a diagonal Gaussian prior for the weights and fix the prior variance to 1. The noise variance of the Gaussian noise model is optimised together with the means and variances of the variational approximation using the variational free energy. We test two different network architectures with the rectified linear activation function, and one and two hidden layers, each of 50 units (100 for the two big datasets), denoted by VI(KW)-1 and VI(KW)-2 respectively.
\item BNN-SGLD: we reuse the same networks with one and two hidden layers as with VI(KW) and approximately sample from the posterior over the weights using Stochastic Gradient Langevin Dynamics (SGLD) \citep{welling2011bayesian}. We place a diagonal Gaussian prior over the weights, and parameterise the observation noise variance as $\sigma^2 = \log (1 + \exp(\kappa))$, a broad Gaussian prior over $\kappa$ and sample $\kappa$ using the same SGLD procedure. Two step sizes, one for the weights and one for $\kappa$, were manually tuned for each dataset. We use Autograd for the implementation of BNN-SGLD and BNN-VI(KW) (\href{github.com/HIPS/autograd}{github.com/HIPS/autograd}).

\item BNN-HMC: We run Hybrid Monte Carlo (HMC) \citep{neal1993bayesian} using the MCMCstuff toolbox \citep{vanhatalo2006mcmc} for networks with one hidden layer. We place a Gaussian prior over the network weights and a broad inverse Gamma hyper-prior for the prior variance. We also assume an inverse Gamma prior over the observation noise variance. The number of leapfrog steps and step size are first tuned using Bayesian optimisation using the pybo package (\href{github.com/mwhoffman/pybo}{github.com/mwhoffman/pybo}). Note that this procedure takes a long time (e.g.~3 days for protein) and the {\it year} dataset is too large to be handled in this way.
\end{itemize}

\begin{figure*}[!t]
	\centering
	\begin{minipage}[b]{.63\textwidth}
		\includegraphics[width=\textwidth]{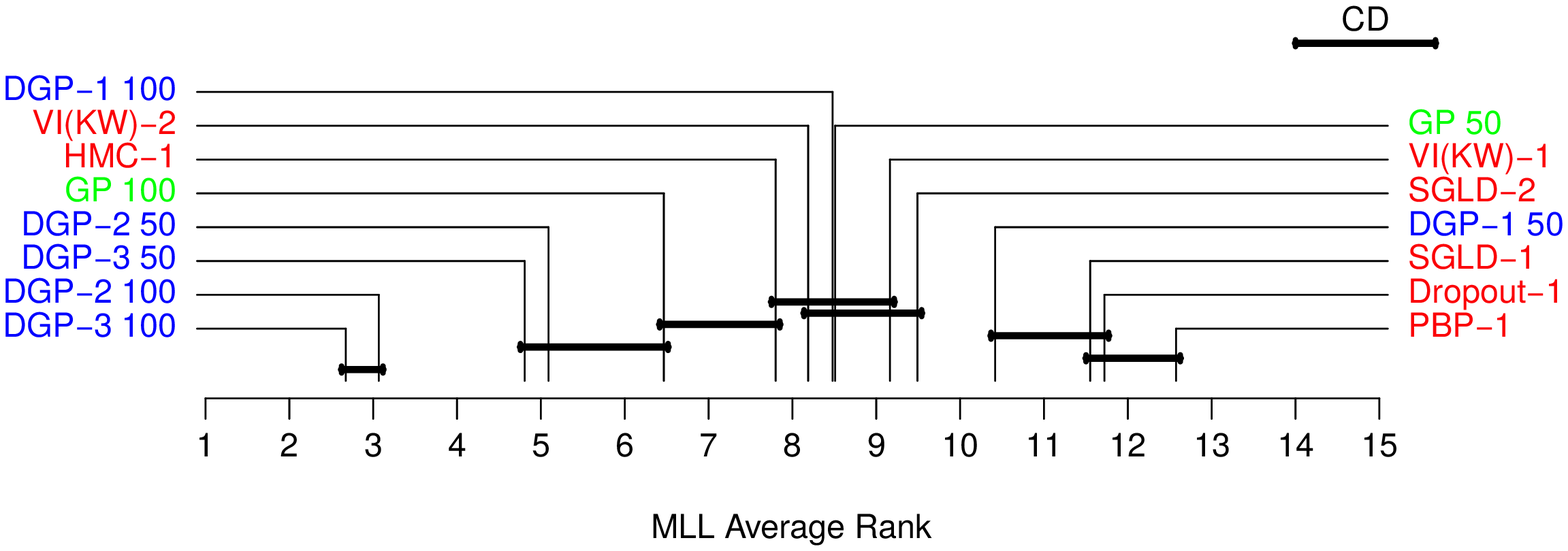}
		\caption{The average rank of all methods across the datasets and their train/test splits, generated based on \citet{demvsar2006statistical}. See the text for more details.}\label{fig:reg_nll_rank}
	\end{minipage}\quad
	\begin{minipage}[b]{.33\textwidth}
		\includegraphics[width=\textwidth]{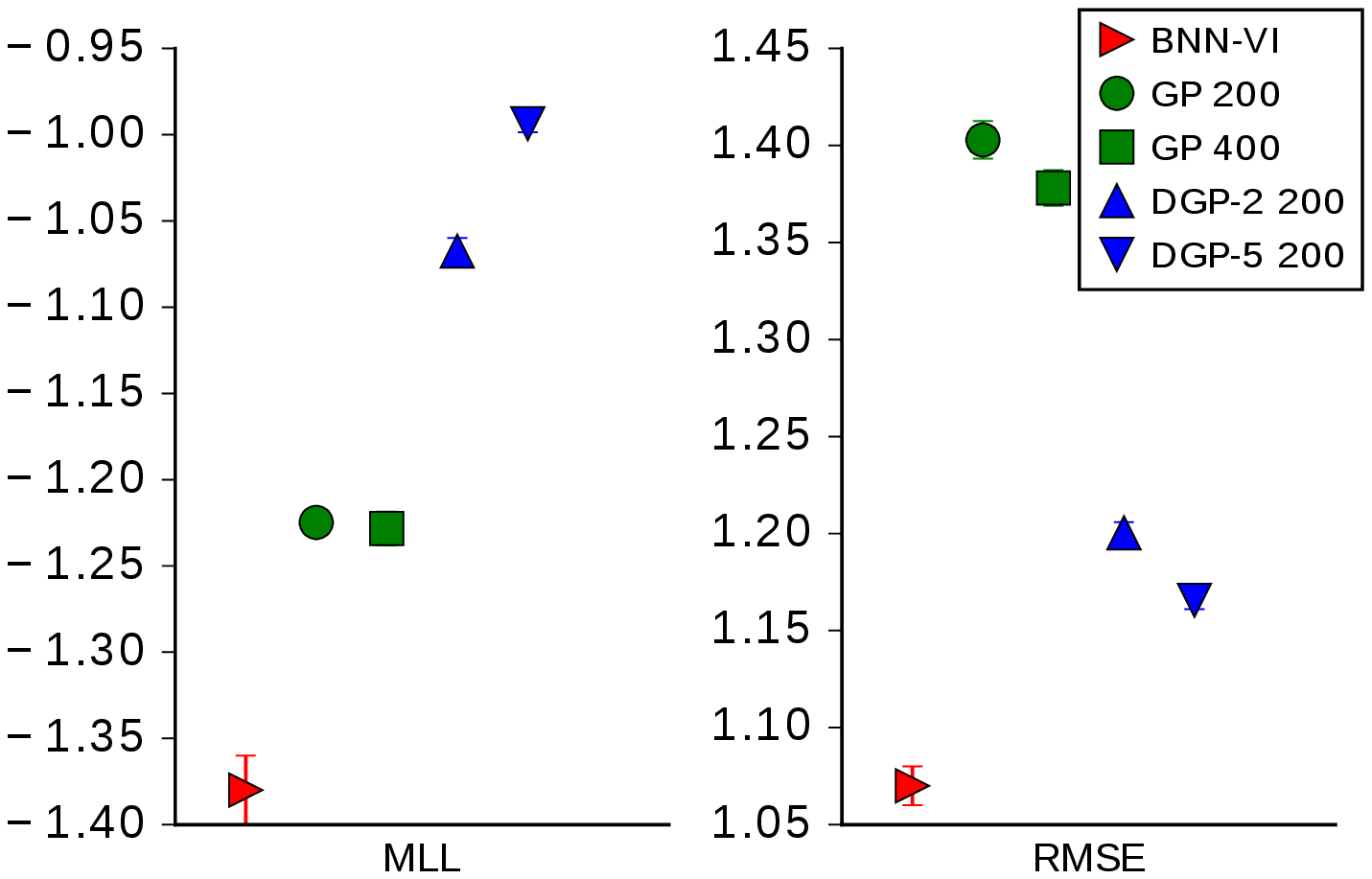}
		\caption{MLL and RMSE results for the photovoltaic molecule regression experiment.}
		\label{fig:molecule_results}
	\end{minipage}
\end{figure*}

%
Figure \ref{fig:reg_results_nll} shows the average test log likelihood (MLL) for a subset of methods with their standard errors. We exclude methods that perform consistently poorly to improve readability. Full results and many more comparisons are included in the supplementary material. We also evaluate the average rank of the MLL performance of all methods across the datasets and their train/test splits and include the results in Figure \ref{fig:reg_nll_rank}. This figure is generated using the comparison scheme provided by \citet{demvsar2006statistical}, and shows statistical differences in the performance of the methods. More precisely, if the gap between the average ranks of any two methods is above the critical distance (shown on the top right), the two methods' performances are statistically significantly different. Methods that are not significantly different from each other are linked by a solid line. The rank result shows that \glspl{dgp} with our inference scheme are the best performing methods overall. Specifically, the DGP-3-100 architecture obtains the best performance on 6 out of 10 datasets and are competitive on the remaining four datasets. The performance of other \gls{dgp} variants follow closely with the exception for DGP-1 which is a standard warped GP, the network with one dimensional hidden layer. DGP-1 performs poorly compared to GP regression, but is still competitive with several methods for BNNs. The results also strongly indicate that the predictive performance is almost always improved by adding extra hidden layers or extra hidden dimensions or extra inducing outputs.

The best non-GP method is BNN-VI(KW)-2 which obtains the best performance on three datasets. However, this method performs poorly on 6 out of 7 remaining datasets, pushing down the corresponding average rank. Despite this, VI(KW) is the best method among all deterministic approximations for BNNs with one or two hidden layers. Overall, the VI approach without the {\it reparameterisation trick} of Graves, Dropout and PBP perform poorly in comparison and give inaccurate predictive uncertainty.

Sampling based methods such as SGLD and HMC obtain good predictive performance overall, but often require more tuning compared to other methods. In particular, HMC appears superior on one dataset, and competitively close to DGP's performance on three other datasets; however, this method does not scale to large datasets.

The results for the RMSE metric follow the same trends with DGP-2 and DGP-3 performing as well or better compared to other methods. Interestingly, BNN-SGLD, despite being ranked relatively low according to the MLL metric, often provides good RMSE results. Full results are included in the supplementary material. 

\subsection{Predicting the efficiency of organic photovoltaic molecules}
\label{exp:molecule}
Having demonstrated the performance of our inference scheme for \glspl{dgp}, we carry out an additional regression experiment on a challenging dataset. We obtain a subset of 60,000 organic molecules and their power conversion efficiency from the Harvard Clean Energy Project (HCEP) (available at \href{http://www.molecularspace.org}{http://www.molecularspace.org}) \citep{hachmann2011harvard}. We use 50,000 molecules for training and 10,000 for testing. The molecules are represented using 512-dimensional binary feature vectors, which were generated using the RDKit package, based on the molecular structures in the canonical SMILES format and a bond radius of 2. The power conversion efficiency of these molecules was estimated using density functional theory, determining whether a molecule could be potentially used as solar cell. The overall aim of the HCEP is to find {\it organic} molecules that are as efficient as their {\it silicon} counterparts. Our aim here is to show \glspl{dgp} are effective predictive models that provide good uncertainty estimates, which can be used for tasks such as Bayesian Optimisation.


We test the method on two \glspl{dgp} with one hidden layer of 2 and 5 dimensions, denoted by DGP-2 and DGP-5 respectively and each \gls{gp} is sparsified using 200 inducing outputs. We compare these against two FITC-GP architectures with 200 and 400 pseudo datapoints respectively. We also repeat the experiment using a Bayesian neural network with two hidden layers, each of 400 hidden units. We use the variational approach with the {\it reparameterisation trick} of \citet{kingma+welling:2014} to perform inference in this model. The noise variance was fixed to 0.16 based on a suggestion in \citet{pyzer2015learning}. Figure \ref{fig:molecule_results} shows the predictive performance by five architectures. The \gls{dgp} with a five dimensional hidden layer significantly outperforms others in terms of test MLL, including the shallow structure with considerably more pseudo datapoints. This result demonstrates the efficacy of \glspl{dgp} in providing good predictive uncertainty estimates, even when the kernel used is a {\it simple} exponentiated quadratic kernel and the input features are binary. Surprisingly, VI(KW), although performing poorly as measured by the MLL, makes good predictions for the mean.

\section{Summary}
This paper has introduced a new and powerful deterministic approximation scheme for \glspl{dgp} based upon an approximate EP algorithm and the FITC approximation to sidestep the computational and analytical intractability. A novel extension of the probabilistic backpropagation algorithm was developed to address a difficult marginalisation problem in the approximate EP algorithm used. The new method was evaluated on eleven datasets and compared against a number of state-of-the-art algorithms for Bayesian neural networks. The results show that the new method for training \glspl{dgp} is superior on 7 out of 11 datasets considered, and performs comparably on the remainder, demonstrating that \glspl{dgp} are a competitive alternative to multi-layer Bayesian neural networks for supervised learning tasks.

The proposed method, in principle, can be applied to classification and unsupervised learning. However, initial work on classification using \glspl{dgp}, as included in the supplementary, does not show a substantial gain over a \gls{gp}. This issue is potentially related to the diagonal Gaussian approximation currently used for the hidden layers from the second layer onwards. A non-diagonal approximation is feasible but more expensive. This can be easily addressed because the computation of our training method can be distributed on GPUs for example, making it even more scalable. We will investigate both problems in future work.

\section*{Acknowledgements}
The authors would like to thank Nilesh Tripuraneni, Alex Matthews, Jes Frellsen and Carl Rasmussen for insightful comments and discussion.
TDB thanks Google for funding his European Doctoral Fellowship.
JMHL acknowledges support from the Rafael del Pino Foundation.
DHL and JMHL acknowledge support from Plan Nacional I+D+i, Grant TIN2013-42351-P, and from CAM, Grant S2013/ICE-2845 CASI-CAM-CM.
YL thanks the Schlumberger Foundation for her Faculty for the Future PhD fellowship. 
RET thanks EPSRC grants EP/G050821/1 and EP/L000776/1.

\bibliography{dgp_aep}

\begin{thebibliography}{34}
\providecommand{\natexlab}[1]{#1}
\providecommand{\url}[1]{\texttt{#1}}
\expandafter\ifx\csname urlstyle\endcsname\relax
  \providecommand{\doi}[1]{doi: #1}\else
  \providecommand{\doi}{doi: \begingroup \urlstyle{rm}\Url}\fi

\bibitem[Barber \& Schottky(1998)Barber and Schottky]{barber-schottky-98b}
Barber, D. and Schottky, B.
\newblock Radial basis functions: a {B}ayesian treatment.
\newblock In \emph{Advances in Neural Information Processing Systems 10}, 1998.

\bibitem[Dai et~al.(2015)Dai, Damianou, Gonz{\'a}lez, and
  Lawrence]{dai2015variational}
Dai, Zhenwen, Damianou, Andreas, Gonz{\'a}lez, Javier, and Lawrence, Neil.
\newblock Variational auto-encoded deep {Gaussian} processes.
\newblock \emph{arXiv preprint arXiv:1511.06455}, 2015.

\bibitem[Damianou(2015)]{andreasthesis}
Damianou, Andreas.
\newblock \emph{Deep {Gaussian} processes and variational propagation of
  uncertainty}.
\newblock PhD thesis, University of Sheffield, 2015.

\bibitem[Damianou \& Lawrence(2013)Damianou and
  Lawrence]{damianou-lawrence:2013a}
Damianou, Andreas~C and Lawrence, Neil~D.
\newblock Deep {G}aussian processes.
\newblock In \emph{16th International Conference on Artificial Intelligence and
  Statistics}, pp.\  207--215, 2013.

\bibitem[Deisenroth \& Mohamed(2012)Deisenroth and Mohamed]{deisenroth2012ep}
Deisenroth, Marc and Mohamed, Shakir.
\newblock Expectation propagation in {G}aussian process dynamical systems.
\newblock In \emph{Advances in Neural Information Processing Systems 25}, pp.\
  2609--2617, 2012.

\bibitem[Dem{\v{s}}ar(2006)]{demvsar2006statistical}
Dem{\v{s}}ar, Janez.
\newblock Statistical comparisons of classifiers over multiple data sets.
\newblock \emph{The Journal of Machine Learning Research}, 7:\penalty0 1--30,
  2006.

\bibitem[Duvenaud et~al.(2014)Duvenaud, Rippel, Adams, and
  Ghahramani]{DuvRipAdaGha14}
Duvenaud, David, Rippel, Oren, Adams, Ryan~P., and Ghahramani, Zoubin.
\newblock Avoiding pathologies in very deep networks.
\newblock In \emph{17th International Conference on Artificial Intelligence and
  Statistics}, 2014.

\bibitem[Gal \& Ghahramani(2015)Gal and Ghahramani]{yarin+zoubin:2015}
Gal, Yarin and Ghahramani, Zoubin.
\newblock Dropout as a {B}ayesian approximation: Representing model uncertainty
  in deep learning.
\newblock \emph{arXiv preprint arXiv:1506.02142}, 2015.

\bibitem[Girard et~al.(2003)Girard, Rasmussen, Qui{\~n}onero-Candela, and
  Murray-Smith]{GirRasQuiMur03}
Girard, Agathe, Rasmussen, Carl~Edward, Qui{\~n}onero-Candela, Joaquin, and
  Murray-Smith, Roderick.
\newblock Gaussian process priors with uncertain inputs --- application to
  multiple-step ahead time series forecasting.
\newblock In \emph{Advances in Neural Information Processing Systems 15}, pp.\
  529--536, 2003.

\bibitem[Graves(2011)]{graves2011practical}
Graves, Alex.
\newblock Practical variational inference for neural networks.
\newblock In \emph{Advances in Neural Information Processing Systems 25}, pp.\
  2348--2356, 2011.

\bibitem[Hachmann et~al.(2011)Hachmann, Olivares-Amaya, Atahan-Evrenk,
  Amador-Bedolla, S{\'a}nchez-Carrera, Gold-Parker, Vogt, Brockway, and
  Aspuru-Guzik]{hachmann2011harvard}
Hachmann, Johannes, Olivares-Amaya, Roberto, Atahan-Evrenk, Sule,
  Amador-Bedolla, Carlos, S{\'a}nchez-Carrera, Roel~S, Gold-Parker, Aryeh,
  Vogt, Leslie, Brockway, Anna~M, and Aspuru-Guzik, Al{\'a}n.
\newblock The {H}arvard clean energy project: large-scale computational
  screening and design of organic photovoltaics on the world community grid.
\newblock \emph{The Journal of Physical Chemistry Letters}, 2\penalty0
  (17):\penalty0 2241--2251, 2011.

\bibitem[Hensman \& Lawrence(2014)Hensman and Lawrence]{hensman+lawrence:2014}
Hensman, James and Lawrence, Neil~D.
\newblock Nested variational compression in deep {G}aussian processes.
\newblock \emph{arXiv preprint arXiv:1412.1370}, 2014.

\bibitem[Hern{\'a}ndez-Lobato \& Adams(2015)Hern{\'a}ndez-Lobato and
  Adams]{hernandez+adams:2015}
Hern{\'a}ndez-Lobato, Jos{\'e}~Miguel and Adams, Ryan~P.
\newblock Probabilistic backpropagation for scalable learning of {B}ayesian
  neural networks.
\newblock In \emph{32nd International Conference on Machine Learning}, 2015.

\bibitem[Kingma \& Ba(2015)Kingma and Ba]{kingma+ba:2015}
Kingma, D.~P. and Ba, J.
\newblock Adam: a method for stochastic optimization.
\newblock In \emph{3rd International Conference on Learning Representations},
  2015.

\bibitem[Kingma \& Welling(2014)Kingma and Welling]{kingma+welling:2014}
Kingma, Diederik~P. and Welling, Max.
\newblock Stochastic gradient {VB} and the variational auto-encoder.
\newblock In \emph{The International Conference on Learning Representations},
  2014.

\bibitem[Lawrence \& Moore(2007)Lawrence and Moore]{lawrence+moore:2007}
Lawrence, Neil~D. and Moore, Andrew~J.
\newblock Hierarchical {G}aussian process latent variable models.
\newblock In \emph{24th International Conference on Machine Learning}, ICML
  '07, pp.\  481--488, New York, NY, USA, 2007.

\bibitem[L{\'a}zaro-Gredilla(2012)]{lazaro2012bayesian}
L{\'a}zaro-Gredilla, Miguel.
\newblock Bayesian warped {G}aussian processes.
\newblock In \emph{Advances in Neural Information Processing Systems 25}, pp.\
  1619--1627, 2012.

\bibitem[Li et~al.(2015)Li, Hern{\'{a}}ndez-Lobato, and Turner]{li+etal:2015}
Li, Yingzhen, Hern{\'{a}}ndez-Lobato, Jos{\'{e}}~Miguel, and Turner, Richard~E.
\newblock Stochastic expectation propagation.
\newblock In \emph{Advances in Neural Information Processing Systems 29}, 2015.

\bibitem[Minka(2001)]{minka2001ep}
Minka, Thomas~P.
\newblock \emph{A family of algorithms for approximate {B}ayesian inference}.
\newblock PhD thesis, Massachusetts Institute of Technology, 2001.

\bibitem[Neal(1993)]{neal1993bayesian}
Neal, Radford~M.
\newblock Bayesian learning via stochastic dynamics.
\newblock In \emph{Advances in Neural Information Processing Systems 6}, pp.\
  475--482, 1993.

\bibitem[Neal(1995)]{neal1995bayesian}
Neal, Radford~M.
\newblock \emph{Bayesian learning for neural networks}.
\newblock PhD thesis, University of Toronto, 1995.

\bibitem[Pyzer-Knapp et~al.(2015)Pyzer-Knapp, Li, and
  Aspuru-Guzik]{pyzer2015learning}
Pyzer-Knapp, Edward~O, Li, Kewei, and Aspuru-Guzik, Alan.
\newblock Learning from the {H}arvard clean energy project: The use of neural
  networks to accelerate materials discovery.
\newblock \emph{Advanced Functional Materials}, 25\penalty0 (41):\penalty0
  6495--6502, 2015.

\bibitem[Qui{\~n}onero-Candela \& Rasmussen(2005)Qui{\~n}onero-Candela and
  Rasmussen]{quinonero+rasmussen:2005}
Qui{\~n}onero-Candela, Joaquin and Rasmussen, Carl~Edward.
\newblock A unifying view of sparse approximate {G}aussian process regression.
\newblock \emph{The Journal of Machine Learning Research}, 6:\penalty0
  1939--1959, 2005.

\bibitem[Rasmussen \& Williams(2005)Rasmussen and Williams]{rasmussen2005gpml}
Rasmussen, Carl~Edward and Williams, Christopher K.~I.
\newblock \emph{{G}aussian Processes for Machine Learning (Adaptive Computation
  and Machine Learning)}.
\newblock The MIT Press, 2005.

\bibitem[Seeger(2007)]{seeger2005ep}
Seeger, Matthias.
\newblock Expectation propagation for exponential families.
\newblock Technical report, Department of EECS, University of California at
  Berkeley, 2007.

\bibitem[Snelson \& Ghahramani(2006)Snelson and
  Ghahramani]{snelson+ghahramani:2006}
Snelson, Edward and Ghahramani, Zoubin.
\newblock Sparse {G}aussian processes using pseudo-inputs.
\newblock In \emph{Advances in Neural Information Processing Systems 19}, pp.\
  1257--1264, 2006.

\bibitem[Snelson et~al.(2004)Snelson, Rasmussen, and Ghahramani]{SneRasGha04}
Snelson, Edward, Rasmussen, Carl~Edward, and Ghahramani, Zoubin.
\newblock Warped {G}aussian processes.
\newblock In \emph{Advances in Neural Information Processing Systems 17}, pp.\
  337--344, Cambridge, MA, USA, 2004.

\bibitem[Sutton \& Barto(1998)Sutton and Barto]{Sutton:1998}
Sutton, Richard~S. and Barto, Andrew~G.
\newblock \emph{Introduction to Reinforcement Learning}.
\newblock MIT Press, Cambridge, MA, USA, 1st edition, 1998.
\newblock ISBN 0262193981.

\bibitem[Titsias(2009)]{titsias2009variational}
Titsias, Michalis~K.
\newblock Variational learning of inducing variables in sparse {G}aussian
  processes.
\newblock In \emph{12th International Conference on Artificial Intelligence and
  Statistics}, pp.\  567--574, 2009.

\bibitem[Titsias \& Lawrence(2010)Titsias and Lawrence]{titsias2010bayesian}
Titsias, Michalis~K and Lawrence, Neil~D.
\newblock Bayesian {G}aussian process latent variable model.
\newblock In \emph{13th International Conference on Artificial Intelligence and
  Statistics}, pp.\  844--851, 2010.

\bibitem[Turner \& Sahani(2011)Turner and Sahani]{turner-and-sahani:2011a}
Turner, R.~E. and Sahani, M.
\newblock Two problems with variational expectation maximisation for
  time-series models.
\newblock In Barber, D., Cemgil, T., and Chiappa, S. (eds.), \emph{Bayesian
  Time series models}, chapter~5, pp.\  109--130. Cambridge University Press,
  2011.

\bibitem[Vanhatalo \& Vehtari(2006)Vanhatalo and Vehtari]{vanhatalo2006mcmc}
Vanhatalo, Jarno and Vehtari, Aki.
\newblock {MCMC} methods for {MLP}-network and {G}aussian process and stuff--a
  documentation for {M}atlab toolbox {MCMCstuff}.
\newblock 2006.
\newblock Laboratory of computational engineering, Helsinki university of
  technology.

\bibitem[Welling \& Teh(2011)Welling and Teh]{welling2011bayesian}
Welling, Max and Teh, Yee~W.
\newblock Bayesian learning via stochastic gradient {L}angevin dynamics.
\newblock In \emph{28th International Conference on Machine Learning}, pp.\
  681--688, 2011.

\bibitem[Wilson \& Adams(2013)Wilson and Adams]{wilson+adams:2013}
Wilson, Andrew and Adams, Ryan.
\newblock Gaussian process kernels for pattern discovery and extrapolation.
\newblock In \emph{30th International Conference on Machine Learning}, pp.\
  1067--1075, 2013.

\end{thebibliography}
\bibliographystyle{icml2016}

\newpage

\begin{appendices}
\section{Extra experimental results}

\subsection{Regression}
\begin{figure*}[!th]
	\centerline{\includegraphics[width=\textwidth]{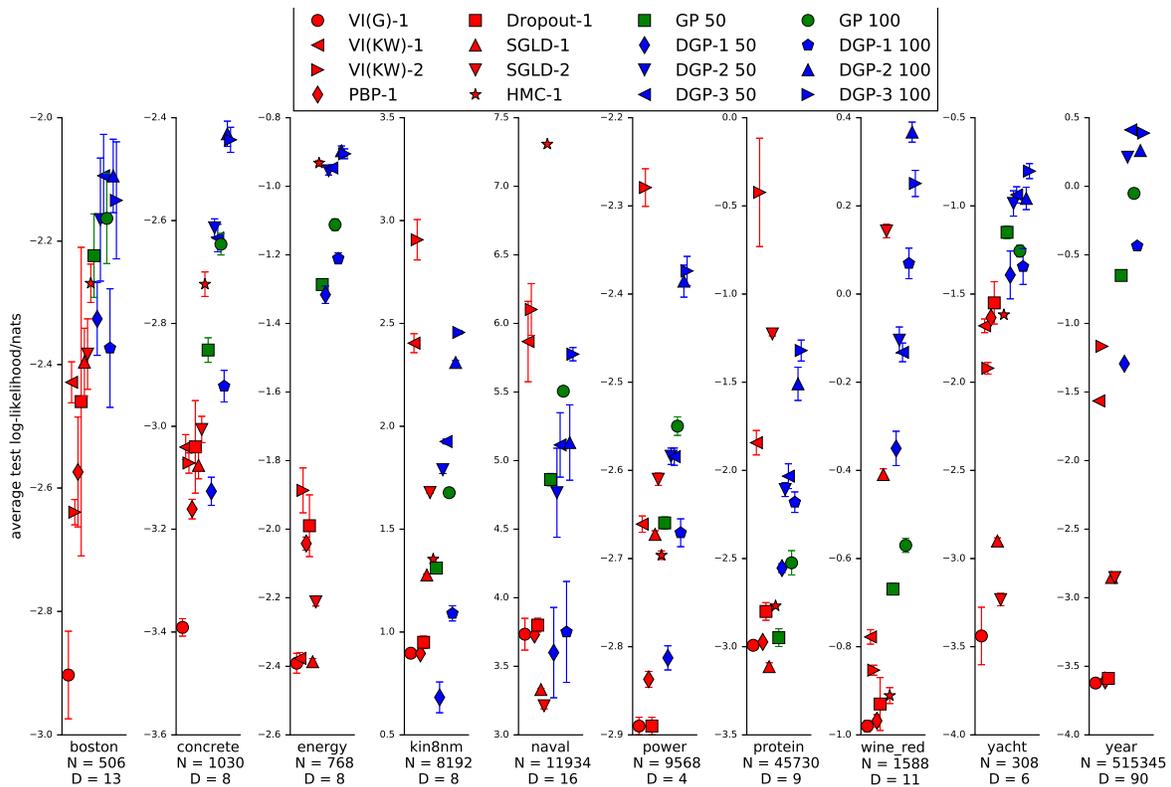}}
	\caption{Average test log likelihood for all methods}
	\label{fig_app:reg_ll_all}
\end{figure*}

\begin{figure*}[!th]
	\centerline{\includegraphics[width=\textwidth]{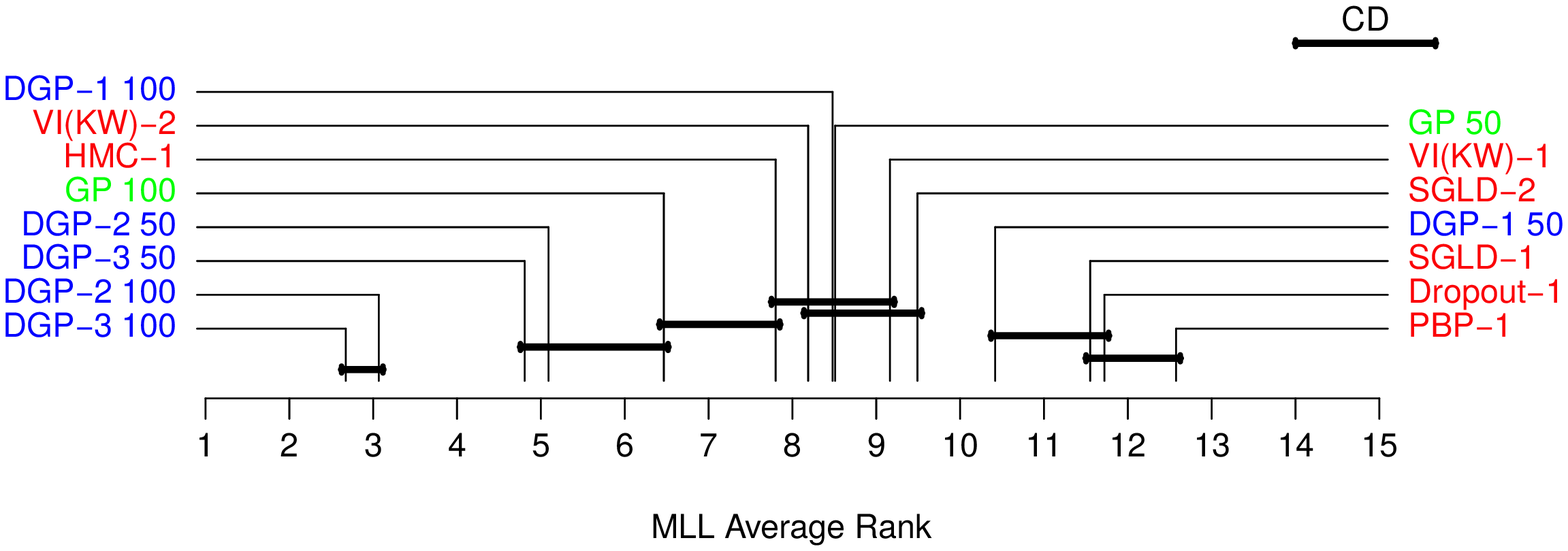}}
	\caption{The average rank based on the test MLL of all methods across the datasets and their train/test splits, generated based on \citet{demvsar2006statistical}. See the main text for more details.}
	\label{fig_app:reg_nll_rank}
\end{figure*}

\begin{figure*}[!th]
	\centerline{\includegraphics[width=\textwidth]{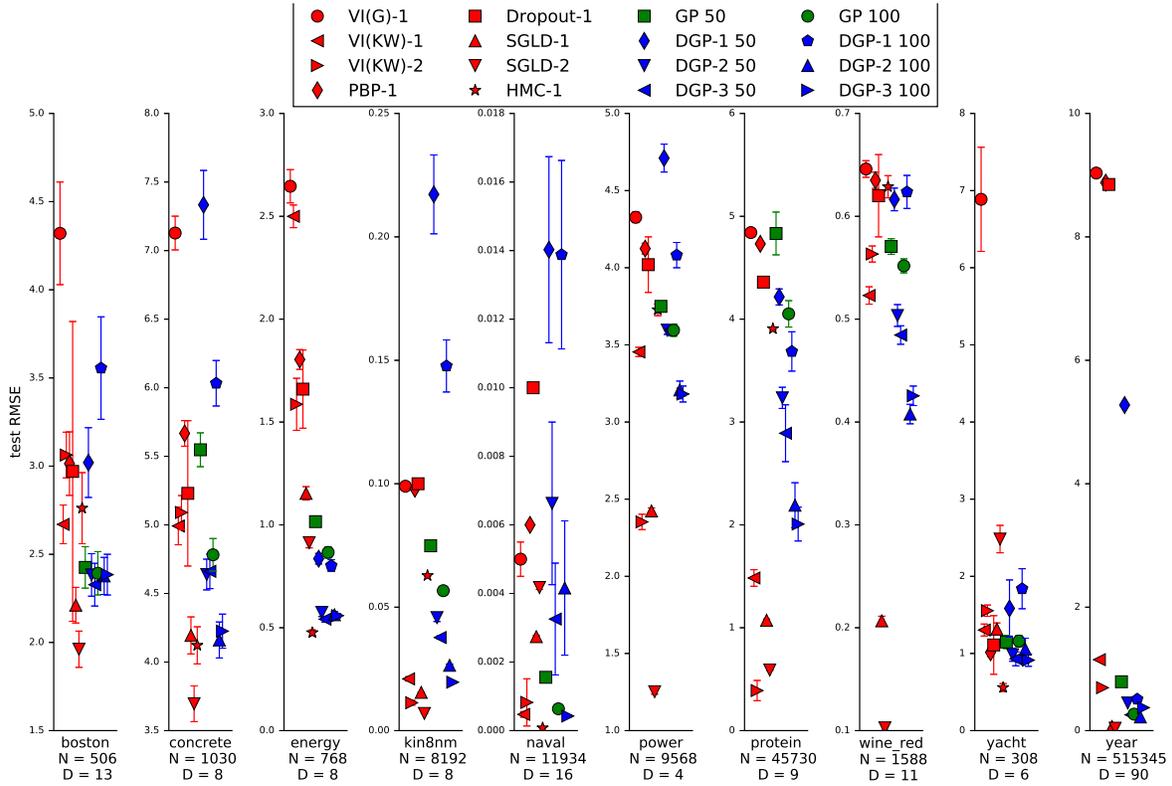}}
	\caption{Average test RMSE for all methods}
	\label{fig_app:reg_rmse_all}
\end{figure*}

\begin{figure*}[!th]
	\centerline{\includegraphics[width=\textwidth]{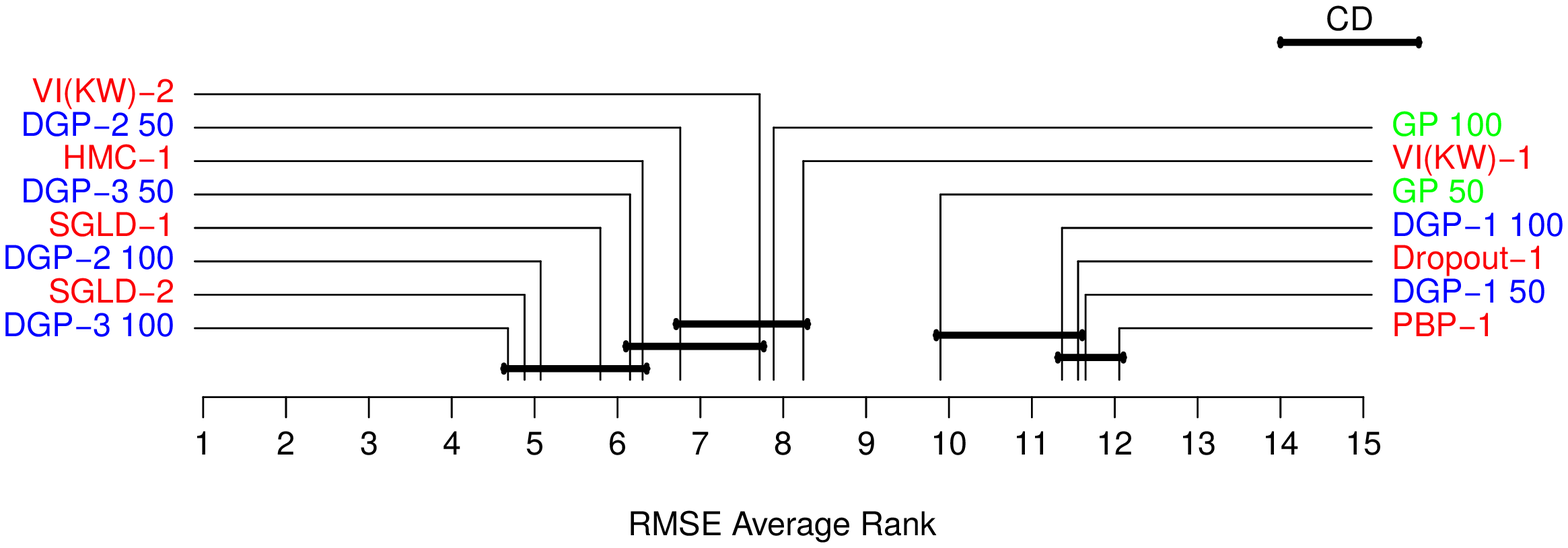}}
	\caption{The average rank based on the test RMSE of all methods across the datasets and their train/test splits, generated based on \citet{demvsar2006statistical}. See the main text for more details.}
	\label{fig_app:reg_rmse_rank}
\end{figure*}

\begin{figure*}[!th]
	\centerline{\includegraphics[width=\textwidth]{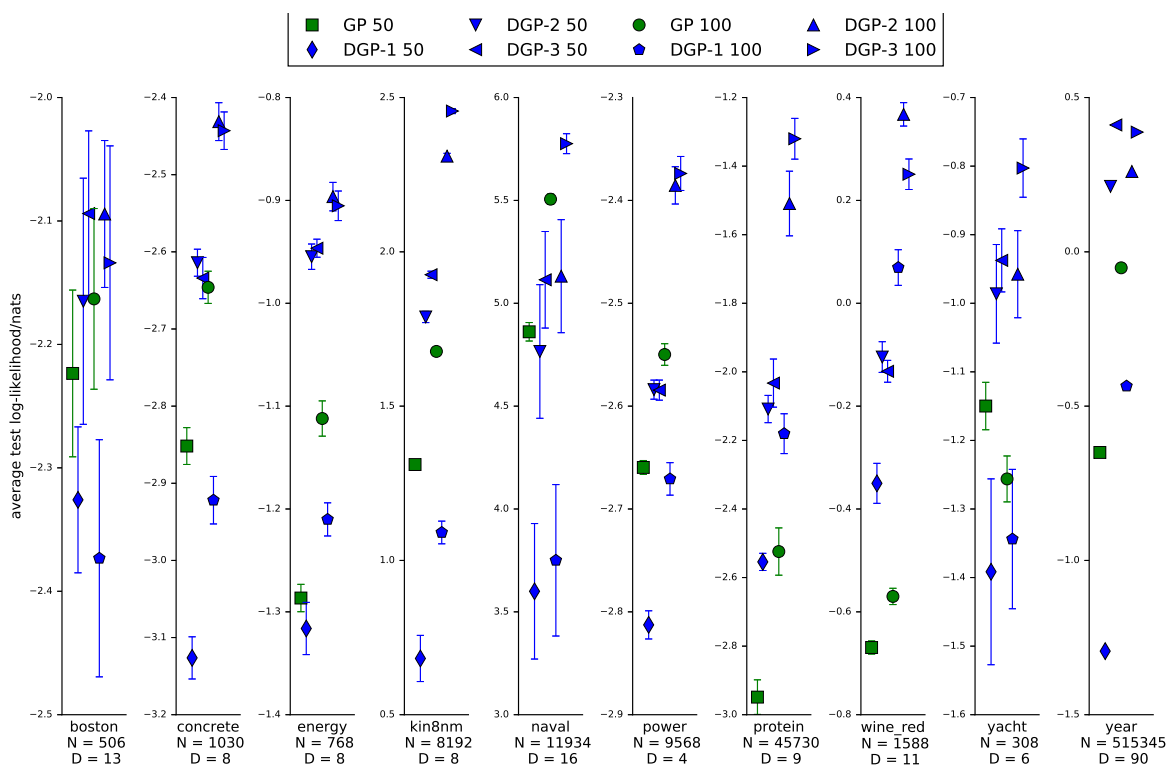}}
	\caption{Average test log likelihood for GP methods}
	\label{fig_app:reg_ll_gp}
\end{figure*}

\begin{figure*}[!th]
	\centerline{\includegraphics[width=\textwidth]{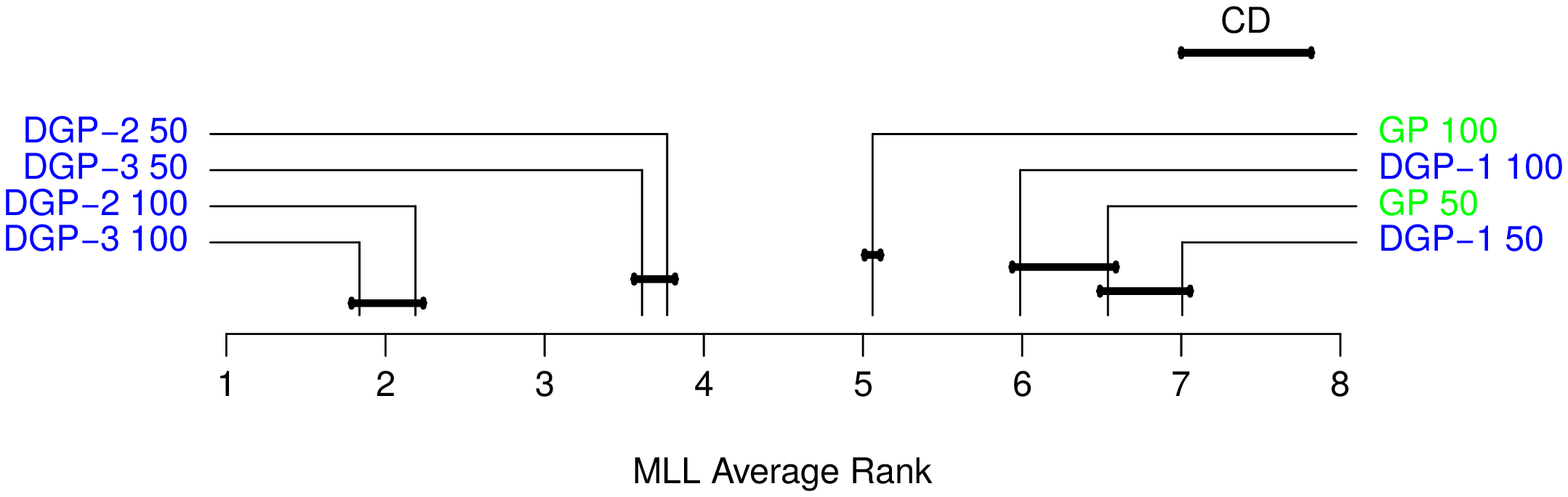}}
	\caption{The average rank based on the test MLL for GP/DGP models across the datasets and their train/test splits, generated based on \citet{demvsar2006statistical}. See the main text for more details.}
	\label{fig_app:reg_nll_rank_gp}
\end{figure*}

\begin{figure*}[!th]
	\centerline{\includegraphics[width=\textwidth]{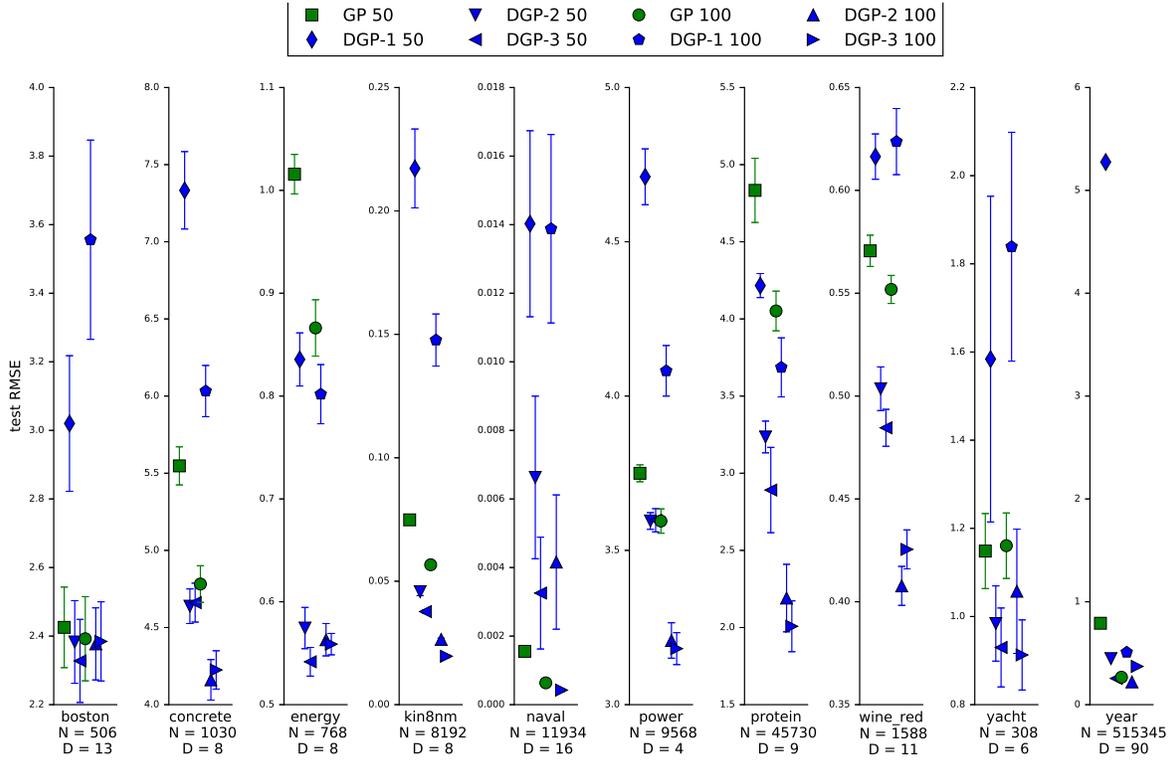}}
	\caption{Average test RMSE for GP methods}
	\label{fig_app:reg_rmse_gp}
\end{figure*}

\begin{figure*}[!th]
	\centerline{\includegraphics[width=\textwidth]{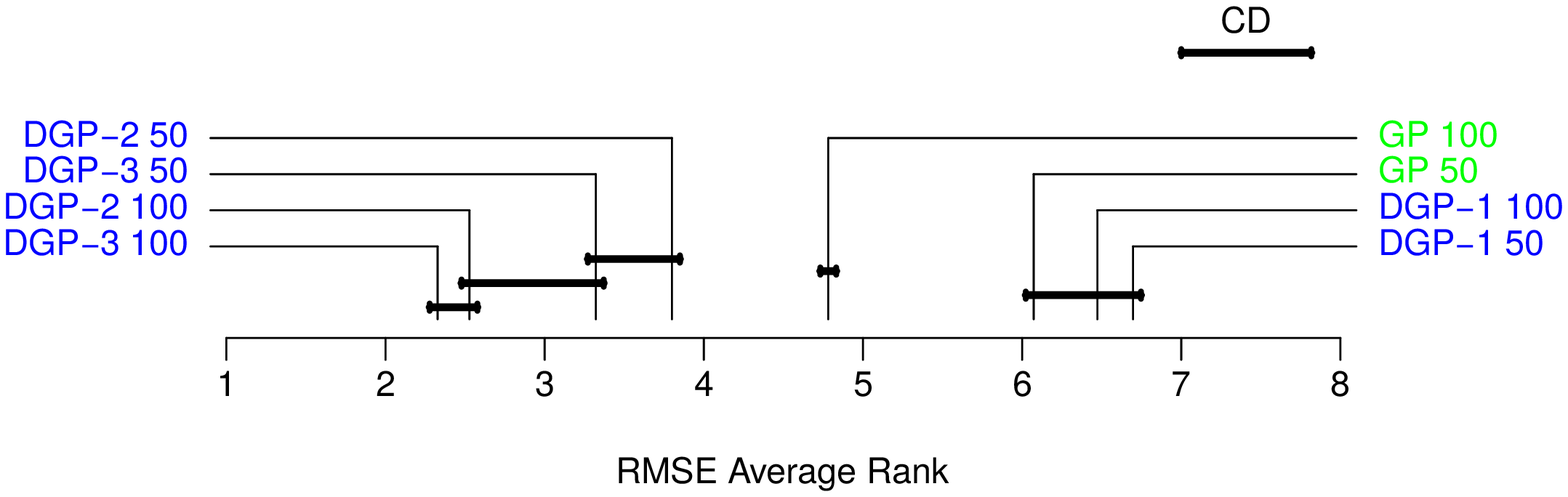}}
	\caption{The average rank based on the test RMSE for GP/DGP models across the datasets and their train/test splits, generated based on \citet{demvsar2006statistical}. See the main text for more details.}
	\label{fig_app:reg_rmse_rank_gp}
\end{figure*}

\begin{figure*}[!th]
	\centerline{\includegraphics[width=\textwidth]{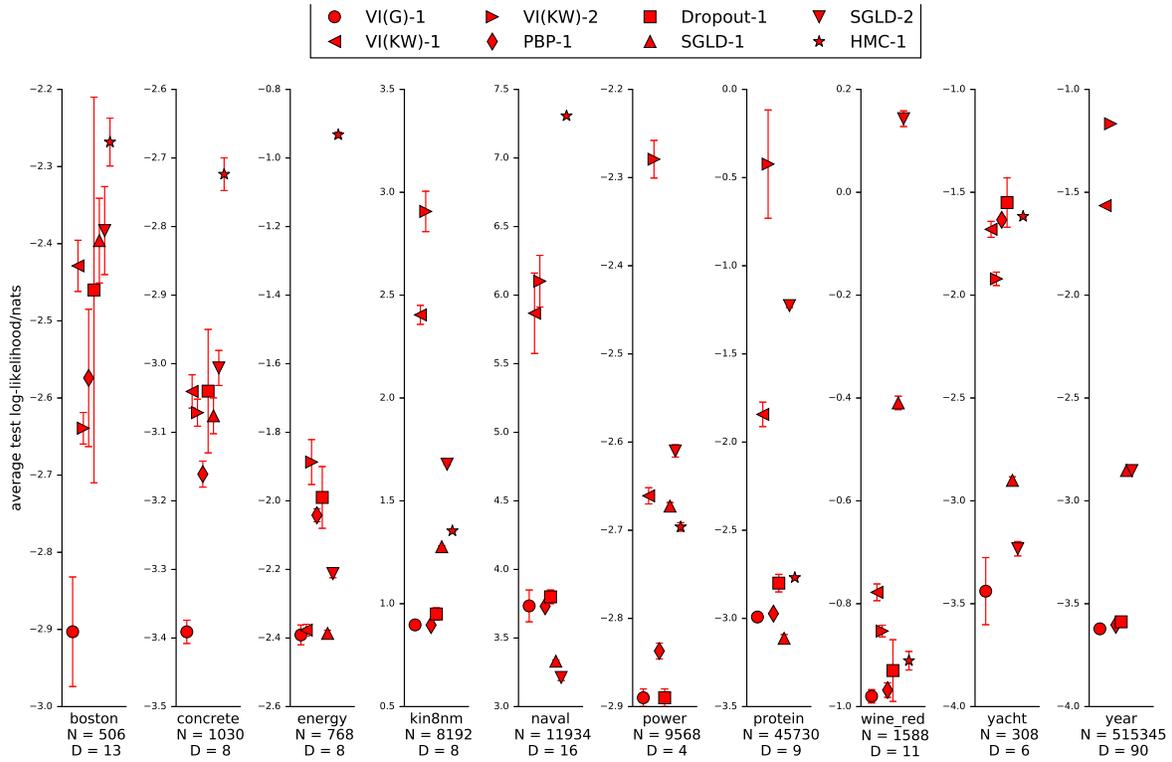}}
	\caption{Average test log likelihood for methods with BNNs}
	\label{fig_app:reg_ll_bnn}
\end{figure*}

\begin{figure*}[!th]
	\centerline{\includegraphics[width=\textwidth]{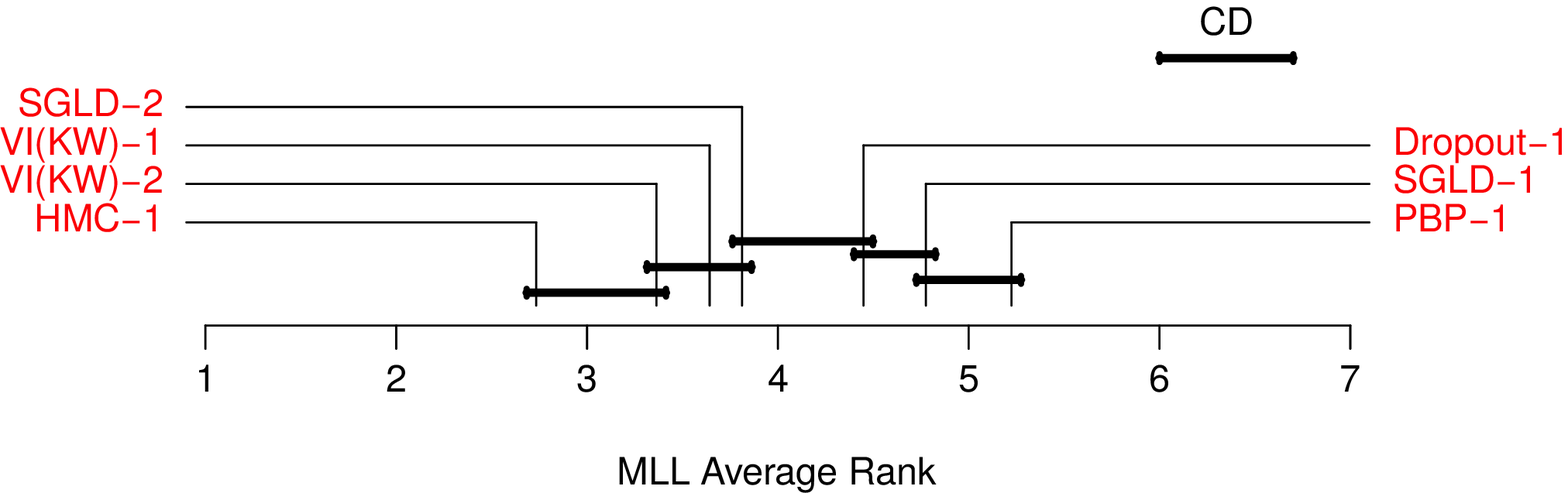}}
	\caption{The average rank based on the test MLL for methods on BNNs across the datasets and their train/test splits, generated based on \citet{demvsar2006statistical}. See the main text for more details.}
	\label{fig_app:reg_nll_rank_bnn}
\end{figure*}

\begin{figure*}[!th]
	\centerline{\includegraphics[width=\textwidth]{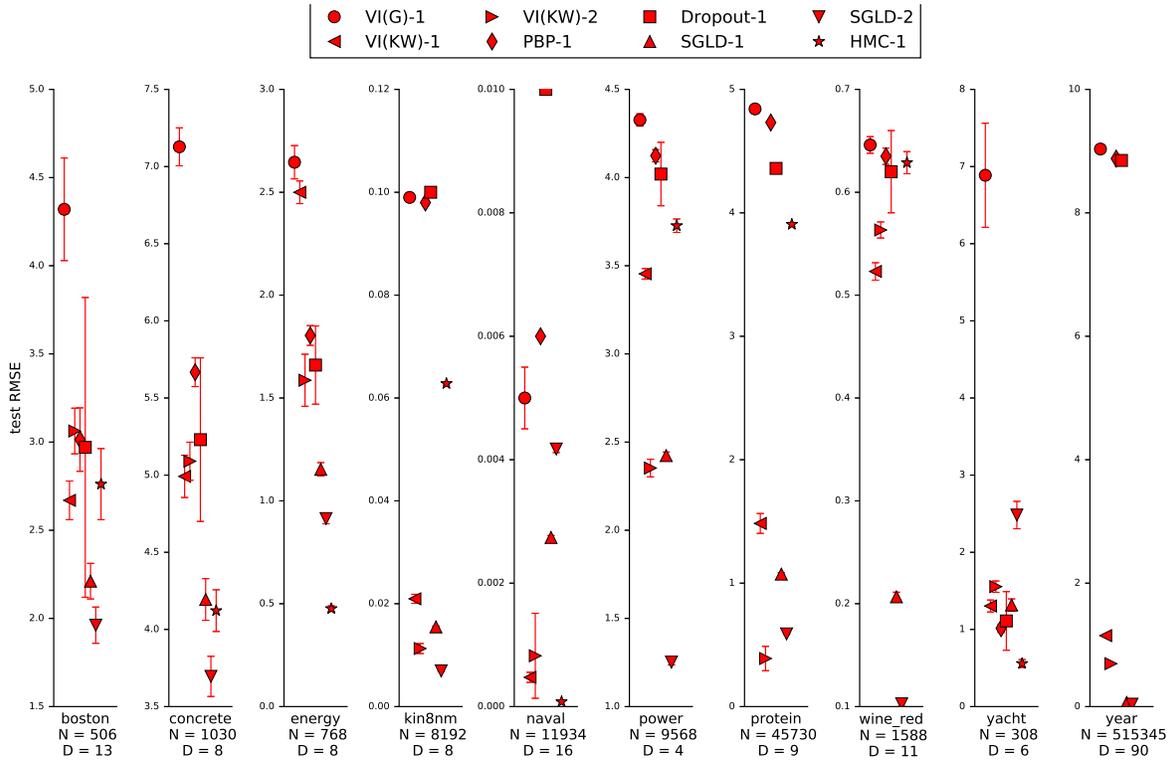}}
	\caption{Average test RMSE for methods with BNNs}
	\label{fig_app:reg_rmse_bnn}
\end{figure*}

\begin{figure*}[!th]
	\centerline{\includegraphics[width=\textwidth]{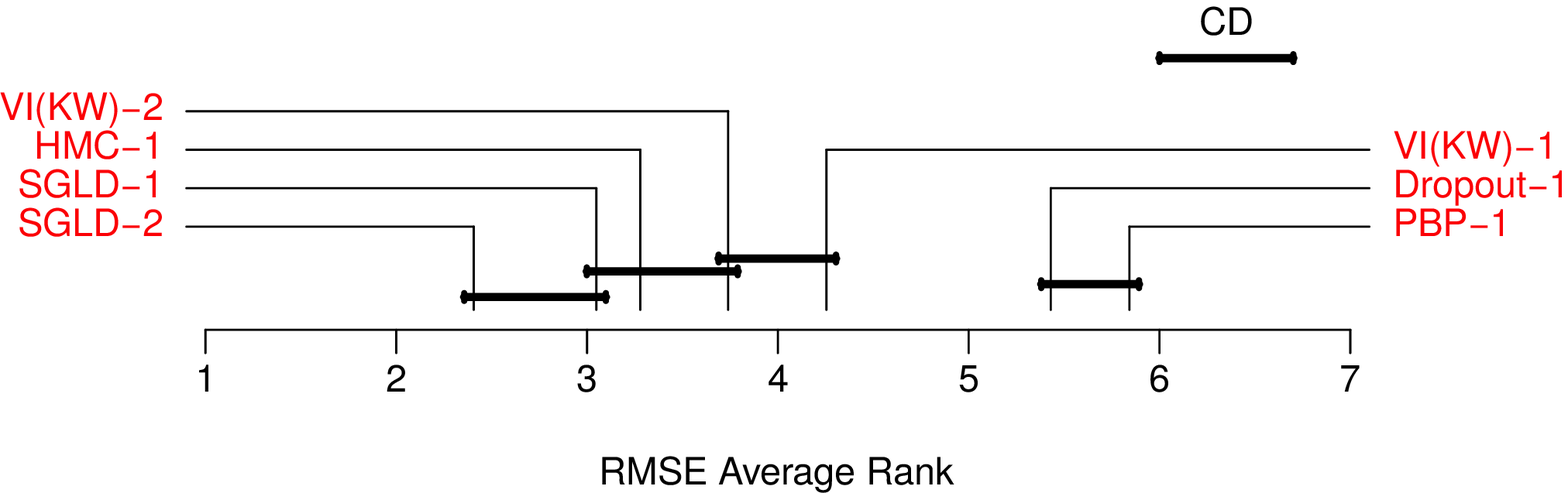}}
	\caption{The average rank based on the test RMSE for methods on BNNs across the datasets and their train/test splits, generated based on \citet{demvsar2006statistical}. See the main text for more details.}
	\label{fig_app:reg_rmse_rank_bnn}
\end{figure*}

Due to the page limitation of the main text, we include here several figures and tables showing the full experimental results and analyses from the regression experiments on 10 UCI datasets. Note that the results for DGPs reported here could be improved further by increasing the number of pseudo datapoints. We choose 50 and 100 pseudo datapoints (or 100 and 200 for the big datasets) so that the training time and prediction time are comparable across all methods. Next we show the full results for the implemented methods and the their average rank across all train/test splits. 
\begin{itemize}
\item Figures \ref{fig_app:reg_ll_all} and \ref{fig_app:reg_nll_rank} show the full MLL results for all methods and all datasets. Part of these results have been included in the main text. These figures show that DGPs with our approximation scheme is superior as measured by the MLL metric, obtaining the top spot in the average ranking table.
\item Figures \ref{fig_app:reg_rmse_all} and \ref{fig_app:reg_rmse_rank} show the full RMSE results for all methods. Surprisingly, though not doing well on the MLL metric, i.e.~providing inaccurate predictive uncertainty, BNN-SGLD with one and two layers are very good at predicting the mean of the test set. DGPs, on average, rival or perform better than this approximate sampling scheme and other methods.
\item Figures \ref{fig_app:reg_ll_gp} and \ref{fig_app:reg_nll_rank_gp} show the subset of the MLL results above, for GP architectures, and their average ranking. This again demonstrate that DGPs are more flexible than GPs, hence always obtain better predictive performance. The only exception is the network with a one dimensional hidden layer or a warped GP which performs poorly relative to other architectures.
\item Similarly, Figures \ref{fig_app:reg_rmse_gp} and \ref{fig_app:reg_rmse_rank_gp} show evidence that increasing the number of layers and hidden dimensions helps improving the accuracy of the predictions.
\item We include a similar analysis for approximate inference methods for BNNs in Figures \ref{fig_app:reg_ll_bnn}, \ref{fig_app:reg_nll_rank_bnn}, \ref{fig_app:reg_rmse_bnn} and \ref{fig_app:reg_rmse_rank_bnn}. This set of results demonstrates that VI(KW) and SGLD with two hidden layers provide good performance on the test sets, outperforming other methods in shallower architectures. HMC with one hidden layer performs well overall, but its running time is much larger compared to other methods. Other deterministic approximations [VI(G), PBP and Dropout] perform poorly overall.
\end{itemize}

Tables \ref{tab:reg_results_ll} and \ref{tab:reg_results_rmse} show the average test log-likelihood and error respectively for all datasets. The best deterministic method for each dataset is bolded, the best method overall (deterministic and sampling) is underlined and emphasised in italic. The average ranks of the methods across the 10 datasets are also included.

\subsection{Binary and multiclass classification}
We test our approximate inference scheme for DGPs with non-Gaussian noise models. However, as shown in Tables \ref{tab:bincla_results_ll} and \ref{tab:multi_results_ll}, DGPs often obtain a marginal gain over GPs, as compared to some substantial improvement in the regression experiments above. We speculate that this is due to our current initialisation strategy and our diagonal Gaussian approximation at last layer for multiclass classification. We will follow this up in future work.
\newcommand{\ica}{\hspace{0.1cm}}
\begin{table}[!ht]
\centering
\caption{Binary cla. experiment: Average test log-likelihood/nats}
\label{tab:bincla_results_ll}
\resizebox{0.6\columnwidth}{!}{%
\begin{tabular}{l@{\ica}r@{$\pm$}l@{\ica}r@{$\pm$}l@{\ica}r@{$\pm$}l@{\ica}r@{$\pm$}l@{\ica}}
\hline
\bf{Dataset}&\multicolumn{2}{c}{\bf{GP D-1}}&\multicolumn{2}{c}{\bf{DGP D-1-1}}&\multicolumn{2}{c}{\bf{DGP D-2-1}}&\multicolumn{2}{c}{\bf{DGP D-3-1}}\\
\hline
australian&\bf{-0.51}&\bf{0.01}&-0.51&0.02&-0.51&0.02&-0.53&0.02\\
breast&-0.05&0.01&-0.04&0.01&\bf{-0.04}&\bf{0.01}&-0.04&0.01\\
crabs&\bf{-0.03}&\bf{0.01}&-0.10&0.05&-0.03&0.01&-0.03&0.01\\
ionoshere&-0.17&0.02&-0.17&0.03&-0.16&0.03&\bf{-0.16}&\bf{0.02}\\
pima&-0.40&0.01&-0.39&0.01&-0.40&0.02&\bf{-0.39}&\bf{0.01}\\
sonar&-0.32&0.03&\bf{-0.29}&\bf{0.03}&-0.30&0.03&-0.31&0.03\\
\hline
\end{tabular}%
}
\end{table}

\begin{table}[!ht]
\centering
\caption{Multiclass experiment: Average test log-likelihood/nats}
\label{tab:multi_results_ll}
\resizebox{0.6\columnwidth}{!}{%
\begin{tabular}{l@{\ica}r@{\ica}r@{\ica}r@{\ica}r@{$\pm$}l@{\ica}r@{$\pm$}l@{\ica}r@{$\pm$}l@{\ica}r@{$\pm$}l@{\ica}}
\hline
\bf{Dataset}&{N}&{D}&{K}&\multicolumn{2}{c}{\bf{GP D-K}}&\multicolumn{2}{c}{\bf{DGP D-1-K}}&\multicolumn{2}{c}{\bf{GP D-2-K}}&\multicolumn{2}{c}{\bf{DGP D-3-K}}\\
\hline
glass&214&9&6&-0.79&0.02&\bf{-0.71}&\bf{0.02}&-0.72&0.02&-0.71&0.02\\
new-thyroid&215&5&3&-0.05&0.01&-0.05&0.01&-0.05&0.02&\bf{-0.04}&\bf{0.01}\\
svmguide2&319&20&3&-0.54&0.02&-0.53&0.02&-0.52&0.02&\bf{-0.51}&\bf{0.02}\\
wine&178&13&3&-0.10&0.01&\bf{-0.07}&\bf{0.01}&-0.07&0.01&-0.07&0.01\\
\hline
\end{tabular}%
}
\end{table}

\begin{landscape}
	
	\begin{table}[ht]
\centering
\tiny
\caption{Regression experiment: Average test log likelihood/nats}
\label{tab:reg_results_ll}
\begin{tabular}{l@{\ica}r@{\ica}r@{\ica}r@{$\pm$}l@{\ica}r@{$\pm$}l@{\ica}r@{$\pm$}l@{\ica}r@{$\pm$}l@{\ica}r@{$\pm$}l@{\ica}r@{$\pm$}l@{\ica}r@{$\pm$}l@{\ica}r@{$\pm$}l@{\ica}r@{$\pm$}l@{\ica}r@{$\pm$}l@{\ica}r@{$\pm$}l@{\ica}r@{$\pm$}l@{\ica}r@{$\pm$}l@{\ica}r@{$\pm$}l@{\ica}r@{$\pm$}l@{\ica}r@{$\pm$}l@{\ica}}
\hline
\bf{Dataset}&{N}&{D}&\multicolumn{2}{c}{\bf{VI(G)-1}}&\multicolumn{2}{c}{\bf{VI(KW)-1}}&\multicolumn{2}{c}{\bf{VI(KW)-2}}&\multicolumn{2}{c}{\bf{PBP-1}}&\multicolumn{2}{c}{\bf{Dropout-1}}&\multicolumn{2}{c}{\bf{SGLD-1}}&\multicolumn{2}{c}{\bf{SGLD-2}}&\multicolumn{2}{c}{\bf{HMC-1}}&\multicolumn{2}{c}{\bf{GP 50}}&\multicolumn{2}{c}{\bf{DGP-1 50}}&\multicolumn{2}{c}{\bf{DGP-2 50}}&\multicolumn{2}{c}{\bf{DGP-3 50}}&\multicolumn{2}{c}{\bf{GP 100}}&\multicolumn{2}{c}{\bf{DGP-1 100}}&\multicolumn{2}{c}{\bf{DGP-2 100}}&\multicolumn{2}{c}{\bf{DGP-3 100}}\\
\hline
boston&506&13&-2.90&0.07&-2.43&0.03&-2.64&0.02&-2.57&0.09&-2.46&0.25&-2.40&0.05&-2.38&0.06&-2.27&0.03&-2.22&0.07&-2.33&0.06&-2.17&0.10&\textbf{\underline{\textit{-2.09}}}&\textbf{\underline{\textit{0.07}}}&-2.16&0.07&-2.37&0.10&-2.09&0.06&-2.13&0.09\\
concrete&1030&8&-3.39&0.02&-3.04&0.02&-3.07&0.02&-3.16&0.02&-3.04&0.09&-3.08&0.03&-3.01&0.03&-2.72&0.02&-2.85&0.02&-3.13&0.03&-2.61&0.02&-2.63&0.03&-2.65&0.02&-2.92&0.03&\textbf{\underline{\textit{-2.43}}}&\textbf{\underline{\textit{0.02}}}&-2.44&0.02\\
energy&768&8&-2.39&0.03&-2.38&0.02&-1.89&0.07&-2.04&0.02&-1.99&0.09&-2.39&0.01&-2.21&0.01&-0.93&0.01&-1.29&0.01&-1.32&0.03&-0.95&0.01&-0.95&0.01&-1.11&0.02&-1.21&0.02&\textbf{\underline{\textit{-0.90}}}&\textbf{\underline{\textit{0.01}}}&-0.91&0.01\\
kin8nm&8192&8&0.90&0.01&2.40&0.05&\textbf{\underline{\textit{2.91}}}&\textbf{\underline{\textit{0.10}}}&0.90&0.01&0.95&0.03&1.28&0.00&1.68&0.00&1.35&0.00&1.31&0.01&0.68&0.07&1.79&0.02&1.93&0.01&1.68&0.01&1.09&0.04&2.31&0.01&2.46&0.01\\
naval&11934&16&3.73&0.12&5.87&0.29&\textbf{6.10}&\textbf{0.19}&3.73&0.01&3.80&0.05&3.33&0.01&3.21&0.02&\underline{\textit{7.31}}&\underline{\textit{0.00}}&4.86&0.04&3.60&0.33&4.77&0.32&5.11&0.23&5.51&0.03&3.75&0.37&5.13&0.27&5.78&0.05\\
power&9568&4&-2.89&0.01&-2.66&0.01&\textbf{\underline{\textit{-2.28}}}&\textbf{\underline{\textit{0.02}}}&-2.84&0.01&-2.89&0.01&-2.67&0.00&-2.61&0.01&-2.70&0.00&-2.66&0.01&-2.81&0.01&-2.58&0.01&-2.58&0.01&-2.55&0.01&-2.67&0.02&-2.39&0.02&-2.37&0.02\\
protein&45730&9&-2.99&0.01&-1.84&0.07&\textbf{\underline{\textit{-0.42}}}&\textbf{\underline{\textit{0.31}}}&-2.97&0.00&-2.80&0.05&-3.11&0.02&-1.23&0.01&-2.77&0.00&-2.95&0.05&-2.55&0.03&-2.11&0.04&-2.03&0.07&-2.52&0.07&-2.18&0.06&-1.51&0.09&-1.32&0.06\\
red wine&1588&11&-0.98&0.01&-0.78&0.02&-0.85&0.01&-0.97&0.01&-0.93&0.06&-0.41&0.01&0.14&0.02&-0.91&0.02&-0.67&0.01&-0.35&0.04&-0.10&0.03&-0.13&0.02&-0.57&0.02&0.07&0.03&\textbf{\underline{\textit{0.37}}}&\textbf{\underline{\textit{0.02}}}&0.25&0.03\\
yacht&308&6&-3.44&0.16&-1.68&0.04&-1.92&0.03&-1.63&0.02&-1.55&0.12&-2.90&0.02&-3.23&0.03&-1.62&0.01&-1.15&0.03&-1.39&0.14&-0.99&0.07&-0.94&0.05&-1.26&0.03&-1.34&0.10&-0.96&0.06&\textbf{\underline{\textit{-0.80}}}&\textbf{\underline{\textit{0.04}}}\\
year&515345&90&-3.62&NA&-1.56&NA&-1.17&NA&-3.60&NA&-3.59&NA&-2.85&NA&-2.85&NA&NA&NA&-0.65&NA&-1.29&NA&0.21&NA&\textbf{\underline{\textit{0.41}}}&NA&-0.05&NA&-0.44&NA&0.26&NA&0.39&NA\\
\hline
\multicolumn{3}{c}{\textbf{Average Rank}}&15.10&0.39&9.00&1.18&7.50&1.70&13.70&0.40&12.10&0.64&12.50&0.75&9.40&1.42&8.80&1.38&8.20&0.69&10.80&0.95&5.30&0.51&4.20&0.66&6.10&0.57&8.20&0.72&2.80&0.49&\textbf{2.30}&\textbf{0.25}\\
\hline
\end{tabular}
\end{table}

	\begin{table}[ht]
\centering
\tiny
\caption{Regression experiment: Test root mean square error}
\label{tab:reg_results_rmse}
\begin{tabular}{l@{\ica}r@{\ica}r@{\ica}r@{$\pm$}l@{\ica}r@{$\pm$}l@{\ica}r@{$\pm$}l@{\ica}r@{$\pm$}l@{\ica}r@{$\pm$}l@{\ica}r@{$\pm$}l@{\ica}r@{$\pm$}l@{\ica}r@{$\pm$}l@{\ica}r@{$\pm$}l@{\ica}r@{$\pm$}l@{\ica}r@{$\pm$}l@{\ica}r@{$\pm$}l@{\ica}r@{$\pm$}l@{\ica}r@{$\pm$}l@{\ica}r@{$\pm$}l@{\ica}r@{$\pm$}l@{\ica}}
\hline
\bf{Dataset}&{N}&{D}&\multicolumn{2}{c}{\bf{VI(G)-1}}&\multicolumn{2}{c}{\bf{VI(KW)-1}}&\multicolumn{2}{c}{\bf{VI(KW)-2}}&\multicolumn{2}{c}{\bf{PBP-1}}&\multicolumn{2}{c}{\bf{Dropout-1}}&\multicolumn{2}{c}{\bf{SGLD-1}}&\multicolumn{2}{c}{\bf{SGLD-2}}&\multicolumn{2}{c}{\bf{HMC-1}}&\multicolumn{2}{c}{\bf{GP 50}}&\multicolumn{2}{c}{\bf{DGP-1 50}}&\multicolumn{2}{c}{\bf{DGP-2 50}}&\multicolumn{2}{c}{\bf{DGP-3 50}}&\multicolumn{2}{c}{\bf{GP 100}}&\multicolumn{2}{c}{\bf{DGP-1 100}}&\multicolumn{2}{c}{\bf{DGP-2 100}}&\multicolumn{2}{c}{\bf{DGP-3 100}}\\
\hline
boston&506&13&4.32&0.29&2.67&0.11&3.06&0.13&3.01&0.18&2.97&0.85&2.21&0.10&\underline{\textit{1.96}}&\underline{\textit{0.10}}&2.76&0.20&2.43&0.12&3.02&0.20&2.38&0.12&\textbf{2.33}&\textbf{0.12}&2.39&0.12&3.56&0.29&2.38&0.11&2.38&0.12\\
concrete&1030&8&7.13&0.12&4.99&0.14&5.09&0.12&5.67&0.09&5.23&0.53&4.19&0.13&\underline{\textit{3.70}}&\underline{\textit{0.13}}&4.12&0.14&5.55&0.12&7.33&0.25&4.64&0.11&4.66&0.13&4.78&0.12&6.03&0.17&\textbf{4.16}&\textbf{0.13}&4.23&0.12\\
energy&768&8&2.65&0.08&2.50&0.06&1.59&0.13&1.80&0.05&1.66&0.19&1.15&0.03&0.91&0.03&\underline{\textit{0.48}}&\underline{\textit{0.01}}&1.02&0.02&0.84&0.03&0.57&0.02&\textbf{0.54}&\textbf{0.01}&0.87&0.03&0.80&0.03&0.56&0.02&0.56&0.01\\
kin8nm&8192&8&0.10&0.00&0.02&0.00&\textbf{0.01}&\textbf{0.00}&0.10&0.00&0.10&0.00&0.02&0.00&\underline{\textit{0.01}}&\underline{\textit{0.00}}&0.06&0.00&0.07&0.00&0.22&0.02&0.05&0.00&0.04&0.00&0.06&0.00&0.15&0.01&0.03&0.00&0.02&0.00\\
naval&11934&16&0.01&0.00&0.00&0.00&0.00&0.00&0.01&0.00&0.01&0.00&0.00&0.00&0.00&0.00&\underline{\textit{0.00}}&\underline{\textit{0.00}}&0.00&0.00&0.01&0.00&0.01&0.00&0.00&0.00&0.00&0.00&0.01&0.00&0.00&0.00&\textbf{0.00}&\textbf{0.00}\\
power&9568&4&4.33&0.04&3.45&0.03&\textbf{2.35}&\textbf{0.05}&4.12&0.03&4.02&0.18&2.42&0.02&\underline{\textit{1.25}}&\underline{\textit{0.02}}&3.73&0.04&3.75&0.03&4.71&0.09&3.60&0.03&3.60&0.04&3.60&0.04&4.08&0.08&3.21&0.06&3.18&0.05\\
protein&45730&9&4.84&0.03&1.48&0.08&\textbf{\underline{\textit{0.39}}}&\textbf{\underline{\textit{0.10}}}&4.73&0.01&4.36&0.04&1.07&0.01&0.59&0.00&3.91&0.02&4.83&0.21&4.22&0.08&3.24&0.10&2.89&0.28&4.05&0.13&3.69&0.19&2.19&0.22&2.01&0.16\\
red wine&1588&11&0.65&0.01&0.52&0.01&0.56&0.01&0.64&0.01&0.62&0.04&0.21&0.00&\underline{\textit{0.10}}&\underline{\textit{0.00}}&0.63&0.01&0.57&0.01&0.62&0.01&0.50&0.01&0.48&0.01&0.55&0.01&0.62&0.02&\textbf{0.41}&\textbf{0.01}&0.43&0.01\\
yacht&308&6&6.89&0.67&1.30&0.08&1.55&0.07&1.01&0.05&1.11&0.38&1.32&0.08&2.48&0.18&\underline{\textit{0.56}}&\underline{\textit{0.05}}&1.15&0.09&1.58&0.37&0.98&0.09&0.93&0.09&1.16&0.07&1.84&0.26&1.06&0.14&\textbf{0.91}&\textbf{0.08}\\
year&515345&90&9.03&NA&1.15&NA&0.70&NA&8.88&NA&8.85&NA&0.07&NA&\underline{\textit{0.04}}&NA&NA&NA&0.79&NA&5.28&NA&0.45&NA&0.26&NA&0.27&NA&0.51&NA&\textbf{0.22}&NA&0.37&NA\\
\hline
\multicolumn{3}{c}{\textbf{Average Rank}}&14.90&0.50&7.90&1.09&7.60&1.42&12.50&0.85&12.00&0.62&4.80&1.08&4.20&1.55&7.50&1.72&10.10&0.74&13.20&0.88&7.00&0.76&5.50&0.72&7.60&0.60&12.20&0.99&4.90&0.57&\textbf{4.10}&\textbf{0.43}\\
\hline
\end{tabular}
\end{table}

\end{landscape}

\section{EP and SEP}

In this section, we summarise the EP and SEP iterative procedures. 
The EP algorithm is often mistaken to be optimising $\mathrm{KL}(p(\uvec|\Xvec,\yvec)||q(\uvec))$; however, this objective function is intractable. 
Instead, EP updates one approximate factor at a time by the following procedure: 1.~remove the factor $\tilde{t}_n(\uvec)$ to form the leave-one-out or cavity distribution $q^{\setminus n}(\uvec) \propto q(\uvec) / \tilde{t}_n(\uvec)$, 2.~minimise $\mathrm{KL}(q^{\setminus n}(\uvec) p(y_n|\uvec,\Xvec_n)||q(\uvec))$, resulting in a new approximate factor $\tilde{t}_n^{\mathrm{new}}(\uvec)$ which can be 3.~combined with the cavity to form the new approximate posterior. 
This procedure is iteratively performed for each datapoint, and often requires several passes through the training set for convergence. One disadvantage of the EP algorithm is the need to store the approximate factors in memory, which costs $\mathcal{O}(NM^2)$.

To sidestep this expensive memory requirement, the SEP algorithm proposes tying the approximate data factors, that is to make some or all factors the same. The simplest case is $q(\uvec) \propto p(\uvec) g(\uvec)^{N}$ where $g(\uvec)$ is the {\it average} data factor. The SEP algorithm, similar to EP, involves iteratively finding the new approximate factor $g_\mathrm{new}(\uvec)$, as follows: 1.~remove the factor $\tilde{g}(\uvec)$ to form the leave-one-out or cavity distribution $q^{\setminus 1}(\uvec) \propto q(\uvec) / \tilde{g}(\uvec)$, 2.~minimise $\mathrm{KL}(q^{\setminus 1}(\uvec) p(y_n|\uvec,\Xvec_n)||q(\uvec))$, resulting in a new approximate factor $\tilde{g}_{\mathrm{new}}(\uvec)$ which can be 3.~combined with the cavity to form the new approximate posterior, and in addition to EP, 4.~perform an explicit update to the {\it average} factor $g(\uvec)$: $g(\uvec) \leftarrow g^{1-\beta}(\uvec)g_\mathrm{new}^{\beta}(\uvec)$, where $\beta$ is a small learning rate.

\section{EP/SEP moment matching step}
We have proposed using the EP approximate marginal likelihood for direct optimisation of the approximate posterior over the pseudo datapoints and the hyperparameters. An alternative is to run SEP/EP to obtain the approximate posterior, and once this is done, obtain the approximate marginal likelihood for hyperparameter tuning and repeat.

As we use Gaussian EP/SEP, the deletion, the update step and the explicit update step in the case of SEP are straightforward. The moment matching step is equivalent to the following updates to the mean and covariance of the approximate posterior:
\begin{align}
\mathbf{m} &= \mathbf{m}^{\setminus 1} + \mathbf{V}^{\setminus 1} \frac{\dd\log \mathcal{Z}}{\dd\mathbf{m}^{\setminus 1}} \nonumber\\
\mathbf{V} &= \mathbf{V}^{\setminus 1} - \mathbf{V}^{\setminus 1} \left[\frac{\dd\log \mathcal{Z}}{\dd\mathbf{m}^{\setminus 1}} \left(\frac{\dd\log \mathcal{Z}}{\dd\mathbf{m}^{\setminus 1}}\right)^{\intercal} -2\frac{\dd\log \mathcal{Z}}{\dd\mathbf{V}^{\setminus 1}}\right]\mathbf{V}^{\setminus 1}, \nonumber
\end{align}
where $q^{\setminus 1}(\uvec) = \norm(\uvec; \mathbf{m}^{\setminus 1}, \mathbf{V}^{\setminus 1})$ is the cavity distribution, obtained by the deletion step.

The inference scheme therefore reduces to evaluating the normalising constant $\mathcal{Z}$ and its gradient. Fortunately, we can approximately compute $\log \mathcal{Z}$ and its gradients using the probabilistic propagation algorithm, in exactly the same way as discussed in the main text.

\section{Computing the gradients of $\log \mathcal{Z}$}

Let $m_l$ and $v_l$ be the mean and variance of the output Gaussian at the $l$-th layer in the forward propagation step, as we have shown in the main text,
\begin{align}
m_{l} &= \psi_{l, 1} \mathbf{A}_l \\ 
v_{l} &= \sigma_l^2 + \psi_{l, 0} + \tr \left( \mathbf{B}_l \psi_{l, 2} \right) - m_{l}^2
\end{align}
where
\begin{myalign}
\psi_{l, 0} &= \mathrm{E}_{q(h_1)} [ K_{h_l,h_l} ]\\
\psi_{l, 1} &= \mathrm{E}_{q(h_{l-1})} [\mathbf{K}_{h_l,\uvec_l}] \\
\psi_{l, 1} &= \mathrm{E}_{q(h_{l-1})} [ \mathbf{K}_{\uvec_l,h_l} \mathbf{K}_{h_l,\uvec_l} ]\\
\mathbf{A}_l &= \mathbf{K}_{\uvec_l,\uvec_l}^{-1} \mathbf{m}_l^{\setminus 1}\\
\mathbf{B}_l &= \mathbf{K}_{\uvec_l,\uvec_l}^{-1} ( \mathbf{V}_l^{\setminus 1} + \mathbf{m}_l^{\setminus 1} \mathbf{m}_l^{\setminus 1, \mathrm{T}} ) \mathbf{K}_{\uvec_l,\uvec_l}^{-1} - \mathbf{K}_{\uvec_l,\uvec_l}^{-1}
\end{myalign}

In the forward propagation step, we need to compute the gradients of $m_l$ and $v_l$ w.r.t.\ $\alpha_l$, the parameters of the model and $m_{l-1}$ and $v_{l-1}$, the mean and variance of the distribution over the input. Let $\beta_l = \{\alpha_l, m_{l-1}, v_{l-1}\}$ As $\mathbf{A}_l$ and $\mathbf{B}_l$ are shared between datapoints, one trick to reduce the computation required for each datapoint is to compute the gradients w.r.t.\ $\mathbf{A}$ and $\mathbf{B}$ first, then combine them at the end of each minibatch. If we assume that $\mathbf{A}_l$ and $\mathbf{B}_l$ are fixed, the gradients of $m_l$ and $v_l$ are as follows
\begin{myalign}
\frac {\dd m_{l}} {\dd \beta_l} &= \frac {\dd \psi_{l, 1}} {\dd \beta_l} \mathbf{A}_l \\ 
\frac {\dd v_{l}} {\dd \beta_l}  &= \frac {\dd \sigma_l^2} {\dd \beta_l} + \frac {\dd \psi_{l, 0}} {\dd \beta_l} + \tr \left( \mathbf{B}_l \frac{ \dd \psi_{l, 2}}  {\dd \beta_l} \right) - 2 m_{l} \frac {\dd m_{l}} {\dd \beta_l} \\
\frac {\dd m_{l}} {\dd \mathbf{A}_l} &= \psi_{l, 1}^\intercal \\
\frac {\dd m_{l}} {\dd \mathbf{B}_l} &= \zero \\
\frac {\dd v_{l}} {\dd \mathbf{A}_l} &= - 2 m_{l} \frac {\dd m_{l}} {\dd \mathbf{A}_l} \\
\frac {\dd v_{l}} {\dd \mathbf{B}_l} &= \psi_{l, 2}^\intercal
\end{myalign}

At the end of the forward step, we can obtain $Z = q(y) = \norm(y; m_L, v_L)$, leading to,
\begin{align}
\log \mathcal{Z} &= - \frac{1}{2} \log (2 \pi v_L) - \frac{1}{2} \frac{(y-m_L)^2}{v_L} \\
\frac{\dd \log \mathcal{Z}} {\dd m_L} &=  \frac{y-m_L}{v_L} \\
\frac{\dd \log \mathcal{Z}} {\dd v_L} &=  - \frac{1}{2 v_L} + \frac{1}{2} \frac{(y-m_L)^2}{v_L^2}.
\end{align}

We are now ready to perform the backpropagation step, that is we compute the gradients of $\log \mathcal{Z}$ w.r.t.\ parameters at a layer $\alpha_l$ using the chain rule,
\begin{align}
\frac{\dd \log \mathcal{Z}} {\dd \alpha_l} &= \frac{\dd \log \mathcal{Z}} {\dd m_l} \frac{\dd m_l} {\dd \alpha_l} + \frac{\dd \log \mathcal{Z}} {\dd v_l} \frac{\dd v_l} {\dd \alpha_l}.
\end{align}
Similarly, we can compute the gradients w.r.t.\ the mean and variance of the input distribution, $m_{l-1}$ and $v_{l-1}$, and $\mathbf{A}_l$ and $\mathbf{B}_l$.

\section{Computing the gradients of the approximate marginal likelihood}
The approximate marginal likelihood as discussed in the main text is as follows,
\begin{myalign} 
\mathcal{F} 
&= -(N-1) \phi(\theta) + N \phi(\theta^{\setminus 1}) - \phi(\theta_{\mathrm{prior}}) + \sum_{n=1}^{N} \log \mathcal{Z}_n \label{eqn:obj} 
\end{myalign}
where $\theta, \theta^{\setminus 1}$ and $\theta_\mathrm{prior}$ are the natural parameters of $q(\uvec)$, $q^{\setminus 1}(\uvec)$ and $p(\uvec)$ respectively, $\phi(\theta)$ is the log normaliser or log partition function of a Gaussian distribution with natural parameters $\theta$ or mean $\mathbf{m}$ and covariance $\mathbf{V}$,
\begin{align}
\phi(\theta) = \frac{1}{2}\log |\mathbf{V}| + \frac{1}{2} \mathbf{m}^\intercal \mathbf{V}^{-1} \mathbf{m},
\end{align}
$\alpha$ is the model hyperameters that we need to tune, and $\log \mathcal{Z}_n = \log \int q^{\setminus n}(\uvec) p(y_n|\uvec,\Xvec_n) \dd \uvec$. Consider the gradient of this objective function w.r.t.\ one parameter $\alpha_i$,
\begin{align}
\frac{\dd \mathcal{F}} {\dd \alpha_i} 
&= -(N-1) \frac{\dd \phi(\theta)} {\dd \alpha_i}  + N \frac{\dd \phi(\theta^{\setminus 1})} {\dd \alpha_i} \nonumber \\
& \quad \quad - \frac{\dd \phi(\theta_{\mathrm{prior}})} {\dd \alpha_i} + \sum_{n=1}^{N} \frac{\dd \log \mathcal{Z}_n} {\dd \alpha_i} \nonumber \\
&= -(N-1) \frac{\dd \phi(\theta)} {\dd \theta} \frac {\dd \theta} {\dd \alpha_i}  + N \frac{\dd \phi(\theta^{\setminus 1})} {\dd \theta^{\setminus 1}} \frac {\dd \theta^{\setminus 1}} {\dd \alpha_i} \nonumber \\ 
& \quad \quad - \frac{\dd \phi(\theta_{\mathrm{prior}})} {\dd \theta_\mathrm{prior}} \frac {\dd \theta_{\mathrm{prior}}} {\dd \alpha_i} + \sum_{n=1}^{N} \frac{\dd \log \mathcal{Z}_n} {\dd \alpha_i} \nonumber \\
&= -(N-1) \eta^\intercal \frac {\dd \theta} {\dd \alpha_i}  + N \eta^{\setminus 1, \intercal} \frac {\dd \theta^{\setminus 1}} {\dd \alpha_i} \nonumber \\ 
& \quad \quad - \eta_{\mathrm{prior}}^\intercal \frac {\dd \theta_{\mathrm{prior}}} {\dd \alpha_i} + \sum_{n=1}^{N} \frac{\dd \log \mathcal{Z}_n} {\dd \alpha_i} \nonumber
\end{align}
where $\eta$, $\eta_{\setminus 1}$ and $\eta_{\mathrm{prior}}$ are the expected sufficient statistics under the $q(\uvec)$, $q^{\setminus 1}(\uvec)$ and $p(\uvec)$ respectively. Specifically, for Gaussian approximate EP as discussed in the main paper, the natural parameters are as follows,
\begin{align}
&q(\uvec):  \theta = \theta_\mathrm{prior} + N \theta_{g} \nonumber\\
&q^{\setminus 1}(\uvec):  \theta^{\setminus 1} = \theta_\mathrm{prior} + (N-1) \theta_{g}\nonumber \\
&p(\uvec):  \theta_\mathrm{prior} \nonumber
\end{align}
leading to
\begin{align}
\frac{\dd \mathcal{F}} {\dd \alpha_i}
&= \left[ -(N-1) \eta^\intercal + N \eta^{\setminus 1, \intercal} - \eta_\mathrm{prior}^{\intercal} \right] \frac {\dd \theta_\mathrm{prior}} {\dd \alpha_i} \nonumber \\
& \quad + N (N-1) \left[ -\eta^\intercal + \eta^{\setminus 1, \intercal} \right] \frac {\dd \theta_g} {\dd \alpha_i} + \sum_{n=1}^{N} \frac{\dd \log \mathcal{Z}_n} {\dd \alpha_i} \nonumber
\end{align}

\section{Dealing with non-Gaussian likelihoods}
In this section, we discuss how to compute the log of $\mathcal{Z} = \int \dd \uvec \, q^{\setminus 1}(\uvec) \, p(y|\uvec,\xvec)$ when we have a non-Gaussian likelihood $p(y|\uvec,\xvec)$. For example, if the observations are binary, we can use the probit likelihood, that is $p(y|\mathrm{f}_L, h_{L-1}) = \mathrm{\phi} (y \mathrm{f}_L)$ where $\phi$ is the Gaussian cdf. We now need to compute,
\begin{align}
\mathcal{Z} 
&= \int q^{\setminus 1}(\uvec)  p(y|\uvec,\xvec) \dd \uvec  \nonumber \\
&= \int q^{\setminus 1}(\uvec) p(\mathrm{f}_L|h_{L-1}, \uvec_L) p(y|\mathrm{f}_L) \dd \uvec \dd h_{L-1} \dd \mathrm{f}_L \nonumber \\
&\approx \int \norm (\mathrm{f}_L; m_\mathrm{f}, v_\mathrm{f}) p(y|\mathrm{f}_L) \dd \mathrm{f}_L\nonumber
\end{align}
where we can find $q(\mathrm{f}_L) = \norm (\mathrm{f}_L; m_\mathrm{f}, v_\mathrm{f})$ using the forward pass of the probabilistic backpropagation. The final integral above can be computed exactly, leading to,
\begin{align}
\mathcal{Z} \approx \mathrm{\phi} \left( \frac{ y m_\mathrm{f}} {\sqrt{v_\mathrm{f} + 1}} \right) \nonumber 
\end{align}

If we have a different likelihood and there is no simple approximation available as above, we can evaluate $\mathcal{Z}$ by Monte Carlo averaging, that is to draw samples from $q(\mathrm{f}_L)$, evaluate the likelihood, then sum and normalise accordingly. However, as we are interested in $\log \mathcal{Z}$ and its gradients, the objective and gradients obtained by Monte Carlo will be slightly biased. This bias is, however, can be significantly reduced by using more samples.

\section{Improving the Gaussian approximation}
In this section, we discuss how to obtain a non-diagonal Gaussian approximation for the hidden variables from the second layer and above, when computing $\log \mathcal{Z}$. Consider a DGP with two GP layer, a one dimensional hidden layer and two dimensional observations $\yvec = [y_1, y_2]$. Following the derivation in the main text, we can exactly marginalise out the inducing outputs for each GP layer:
\begin{align}
\mathcal{Z} = \int \dd h_1 q(\yvec|h_1) q(h_1) \label{eqn:Z}
\end{align}
where $q(h_1) = \norm(h_1; m_1, v_1)$ and 
\begin{align}
q(\yvec|h_1) &= \norm (\yvec|h_1; \mathbf{m}_{\yvec|h_1}, \mathbf{V}_{\yvec|h_1})\nonumber\\
&= \norm \left(\yvec|h_1; \begin{bmatrix} m_{y_1|h_1} \\ m_{y_2|h_1} \end{bmatrix} ,  \begin{bmatrix} v_{y_1|h_1} & 0 \\ 0 & v_{y_2|h_1} \end{bmatrix} \right) \nonumber
\end{align}
since we assume that there are two independent GPs in the second layer, and the distribution above is a conditional given the input to the second layer, $h_1$. Importantly, we need to integrate out $h_1$ in eqn.~\eqref{eqn:Z}. As such, the resulting distribution over $\yvec$ become a complicated distribution in which $y_1$ and $y_2$ are strongly correlated. Consequently, any approximation that breaks this dependency could be poor. We aim to approximate this distribution by a non-diagonal Gaussian with the same moments, that is in words, the approximating Gaussian will have the mean being the expected mean, and the new covariance being the expected covariance plus the covariance of the mean,
\begin{align}
\mathbf{m}_{\yvec} &= \mathrm{E}_{q(h_1)} [\mathbf{m}_{\yvec|h_1}] \\
\mathbf{V}_{\yvec} &= \mathrm{E}_{q(h_1)} [\mathbf{V}_{\yvec|h_1}]  + \mathrm{covar}_{q(h_1)}[\mathbf{m}_{\yvec|h_1}]
\end{align}
Substitute the mean and covariance of the conditional $q(\yvec|h_1)$ into the above expressions gives us,
\begin{align}
\mathbf{m}_{\yvec} &= \begin{bmatrix} \mathrm{E}_{q(h_1)} [m_{y_1|h_1}] \\ \mathrm{E}_{q(h_1)} [m_{y_2|h_1}]\end{bmatrix}
\end{align}
and 
\begin{align}
\mathbf{V}_{\yvec} 
&= \begin{bmatrix} \mathrm{E}_{q(h_1)} [v_{y_1|h_1}] & 0 \\ 0 & \mathrm{E}_{q(h_1)} [v_{y_2|h_1}]\end{bmatrix} \nonumber \\ 
&\quad + \begin{bmatrix} \mathrm{E}_{q(h_1)} [m^2_{y_1|h_1}] & \mathrm{E}_{q(h_1)} [m_{y_1|h_1} m_{y_2|h_1}] \\ \mathrm{E}_{q(h_1)} [m_{y_1|h_1} m_{y_2|h_1}] & \mathrm{E}_{q(h_1)} [m^2_{y_2|h_1}]\end{bmatrix} \nonumber \\
&\quad - \mathbf{m}_{\yvec} \mathbf{m}_{\yvec}^\intercal
\end{align}

Note that the diagonal elements of $\mathbf{V}_{\yvec}$ are identical to the expression for the variance in the main text for the single dimensional case.
\end{appendices}

\end{document}